\documentclass[10pt,twocolumn,letterpaper]{article}
\usepackage[T1]{fontenc}   %
\usepackage[pagenumbers]{cvpr}

\renewcommand{\paragraph}[1]{\vspace{.5em}\noindent\textbf{#1.}}

\usepackage{booktabs,multirow,amsmath,amssymb,graphicx,xspace}
\usepackage{colortbl}
\definecolor{cvprblue}{rgb}{0.21,0.49,0.74}
\definecolor{oursbg}{RGB}{253,226,204}   %
\usepackage[pagebackref,breaklinks,colorlinks,allcolors=cvprblue]{hyperref}
\AtBeginDocument{\crefname{appendix}{Appendix}{Appendices}\Crefname{appendix}{Appendix}{Appendices}\crefname{subappendix}{Appendix}{Appendices}\Crefname{subappendix}{Appendix}{Appendices}}
\title{Teaching Time Without Losing General Retrieval:\\ Adapting Multimodal Embeddings for Time-Sensitive Video Retrieval}
\author{Toya Oyama$^{1,2}$ \qquad Rainer Lienhart$^{3}$ \qquad Shin'ichi Satoh$^{2,1}$\\
$^{1}$The University of Tokyo \qquad $^{2}$National Institute of Informatics \qquad $^{3}$University of Augsburg}
\begin{document}
\maketitle
\begin{abstract}
In text-to-video retrieval, MLLM-based embedding models do well when the text describes the scene (general retrieval), but poorly when it describes actions and their direction (time-sensitive retrieval).
TARA, a pioneering work, fine-tunes such a model on text triplets with temporal hard negatives; it improves time-sensitive retrieval but lowers general retrieval (mean R@1 over six benchmarks in our rerun on Qwen3-VL-Embedding-8B: $58.4 \rightarrow 44.8$), and its released model keeps general retrieval only when ensembled with Qwen3-VL-Embedding.
Our goal is strong general and time-sensitive retrieval with one model and one vector per clip. Three findings lead to three simple components.
\emph{Keep loss:} general retrieval follows how far the output embeddings drift from the base model, not how far the weights move; we therefore keep them near the base model, which keeps most general retrieval even without general-domain data.
\emph{Pair loss:} the text embeddings of an action description and its reversed-direction version are well separated, but the video embeddings of a clip and its reversed copy are close; we therefore widen the gap between the video embeddings of each clip and its reversed copy, in the direction of the difference between their two text embeddings.
\emph{Middle-layer LoRA:} frames are combined in the middle layers, and temporal information peaks in or just after them; we therefore train LoRA only on these layers, which nearly matches training all weights on most time-sensitive benchmarks with $0.15\%$ of the parameters and keeps general retrieval better.
General R@1 stays at $58.4$, while R@1 with each clip's reversed copy in the gallery rises from $55.5$ to $63.2$; the gains carry over to unseen benchmarks and InternVL2-8B-RA.
The keep loss is also a remedy for the TARA recipe (general R@1: $44.8 \rightarrow 57.8$), and ensembling TARA with Ours gives the highest scores on RTime and unseen CiA among the compared models.
\end{abstract}

\begin{figure}[t]
  \centering
  \includegraphics[width=.92\linewidth]{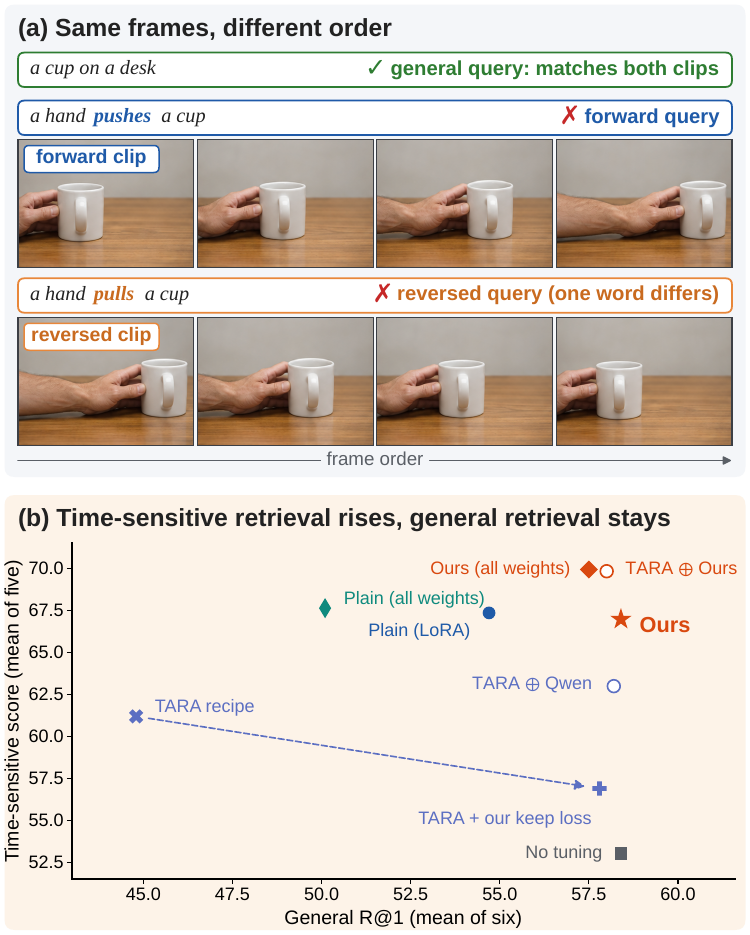}
  \caption{\textbf{Training for time-sensitive retrieval usually lowers general retrieval; Ours gains without this drop.}
  (a)~The forward and reversed clips show the same frames in reverse order, and their text queries differ only in one word; the base model barely separates such clips (mean cosine $0.95$ on RTime).
  (b)~General R@1 (mean of six) against the mean of five time-sensitive scores (RTime Hard and Binary, SSv2-T, and unseen CiA-Charades and CiA-EPIC Chiral; \cref{sec:observations}), three seeds. Plain: no keep or pair loss. Ours and Plain (LoRA) use LoRA on the middle layers. The arrow adds our keep loss to our rerun of the TARA recipe. $\oplus$: the released TARA model ensembled with the base model or Ours (\cref{tab:compare}).}
  \label{fig:teaser}
\end{figure}

\section{Introduction}
\label{sec:intro}
Embedding models built on multimodal large language models (MLLMs) retrieve videos well but are weak when the text query is about what happens over time: when the direction of a motion decides the action.
On Qwen3-VL-Embedding-8B~\cite{li2026qwen3vlemb} (Qwen), Recall@1 (R@1) is $56.1$ on MSR-VTT~\cite{xu2016msrvtt}, whose queries mostly describe what is visible, but only $22.4$ on SSv2-Template~\cite{lei2023singularity}, whose queries describe actions.
On RTime~\cite{du2024rtime}, R@1 falls from $89.5$ to $55.5$ when the reversed copy of each clip (see \cref{fig:teaser}a for an example) is added to the gallery. We call this RTime Hard.

Training for time-sensitive retrieval improves it but lowers general retrieval performance.
With the recipe of TARA~\cite{bagad2026tara}, which fine-tunes on text triplets with temporal hard negatives, the mean general R@1 of Qwen falls from $58.4$ to $44.8$ (\cref{sec:experiments}). The released TARA model keeps general retrieval only when ensembled with Qwen, which needs a second embedding model and a vector twice as long.
Plain training (Plain: contrastive losses on time-sensitive data, with no constraint on the embeddings) also lowers general R@1, to $54.7$ with low-rank adapters (LoRA)~\cite{hu2022lora} and to $50.1$ with all weights of the language model (\cref{fig:teaser}b).

We ask whether a model with one vector per clip can improve on time-sensitive queries while preserving general R@1 performance.
Three observations (\cref{sec:observations}) suggest three hypotheses.
First, Plain training moves the clip embeddings away from those of the base model and breaks their neighborhoods. (H1)~Holding the output embeddings near those of the base model keeps general retrieval.
Second, the text embeddings of an action description and its reversed-direction version are well separated, but the video embeddings of a clip and its reversed copy are close. (H2)~Pushing each clip away from its reversed copy raises time-sensitive retrieval.
Third, layer-wise probes and attention blocking show that frames are combined in the middle layers. Temporal information peaks in or just after them (L15--L28 on Qwen) and then fades. (H3)~Training these layers gains more than training only the later layers.

Our recipe (Ours) follows the hypotheses. Starting from Plain training, we add a keep loss on the output embeddings (H1) and a pair-margin loss on each clip and its own reversed copy (H2), which pushes the difference of their video embeddings toward the difference of their text embeddings. We train LoRA on the middle layers (H3).
Each part is known: the keep loss is an output anchor, as in LwF~\cite{li2016lwf} and BCT~\cite{shen2020bct}, and reversed clips serve as negatives in earlier work~\cite{bagad2023testoftime,du2024rtime}. Our contribution is the analysis that says which parts are needed and why. Interpolating the weights of Ours without the keep loss toward the base model (WiSE-FT~\cite{wortsman2022wise}) reaches about the same point as Ours, as H1 predicts. The keep loss achieves this in a single run, without choosing the interpolation weight.

With one vector per clip, Ours keeps general R@1 at $58.4$, while RTime Hard rises from $55.5$ to $63.2$ and retrieval of opposite actions on two unseen benchmarks from Chirality in Action (CiA)~\cite{bagad2025chirality}, CiA-Charades and CiA-EPIC, gains $11$ and $13$ points, mostly through frame order. The results hold on InternVL2-8B-RA.

Our contributions are:
\begin{itemize}
  \item \emph{Output drift (H1).} Across $28$ variants, general R@1 tracks how close the output embeddings stay to those of the base model, not how far the weights move. This holds whether a loss keeps them close during training or weight interpolation restores them after training. Keeping the weights near their initial values during training (L2-SP, a lower learning rate) preserves less general R@1. With a keep loss, Ours keeps most general retrieval even without general-domain data and lies above WiSE-FT of Plain training, and the loss restores general R@1 in the TARA recipe (\cref{sec:observations,sec:experiments,sec:analysis}).
  \item \emph{Direction gap (H2).} The video embeddings of a clip and its reversed copy are close, while the text embeddings of their descriptions are well separated. A pair-margin loss that widens the gap between these two video embeddings toward the difference of their text embeddings raises time-sensitive retrieval on two models, including opposite actions on unseen benchmarks (\cref{sec:observations,sec:method,sec:experiments}).
  \item \emph{Middle-layer peak (H3).} Temporal information peaks in or just after the middle layers, where frames are combined, and then fades, unlike appearance. LoRA on these layers, with $0.15\%$ of the parameters, raises RTime and unseen CiA nearly as much as all weights and keeps general R@1 better (\cref{sec:observations,sec:experiments}).
\end{itemize}

\section{Related Work}
\label{sec:related}
\paragraph{MLLM-based embedding models}
Many works turn MLLMs into one-vector embedders, trained on text~\cite{jiang2024e5v,tzachor2026vidvec}, text and images~\cite{jiang2025vlm2vec,liu2025lamra}, captioning then text~\cite{xu2026carebench} or large multimodal sets~\cite{meng2025vlm2vecv2,kong2025unite,guo2025gve,gupta2026ville,li2026qwen3vlemb}. MARS~\cite{jung2026mars} keeps several vectors.
These models target general retrieval. We adapt such a model to time-sensitive retrieval.

\paragraph{Time-sensitive retrieval}
Single frames solve many video-language tasks~\cite{lei2023singularity}, so new benchmarks test the order of events and the direction of motion~\cite{lei2023singularity,du2024rtime,bagad2025chirality,li2024vitatecs,kim2025videocomp}.
Training methods add temporal hard negatives such as reordered or reversed clips~\cite{bagad2023testoftime,wang2023paxion,du2024rtime,tv2026vitalx} or temporal antonym captions in TARA~\cite{bagad2026tara}, which is trained on text data alone.
We also use reversed clips and find that a loss pairing each clip with its own reversed copy gains more than stronger hard negatives.
On a strong embedder, the released TARA model keeps general R@1 only when ensembled with Qwen3-VL-Embedding, and ViTAL-X~\cite{tv2026vitalx} starts from a frozen SigLIP 2. We fine-tune a strong embedder and keep its general R@1 with one model and one vector per clip.

\paragraph{Temporal information inside the model}
Studies of video LLMs for question answering find that temporal reasoning starts in the early to middle layers~\cite{kim2026mapflow,shi2026causality} and that temporal information fades near the output~\cite{tv2026reveal,shin2026fades}. Several methods recover it~\cite{shin2026fades,lin2026spod,fu2026merit,han2026tracing,xue2025arrowrl}. Blocking attention between frames in all layers, however, costs little question-answering accuracy~\cite{gou2026empirical}.
In retrieval, a middle layer of a video MLLM can retrieve better than the final layer without training~\cite{tzachor2026vidvec}, and MLLM and CLIP retrievers gain from fusing features of several layers~\cite{jung2026mars,liu2023stan}. These works fuse or select layers but do not ask what those layers encode. We measure temporal information layer by layer and then train the middle layers, keeping one vector from the final layer.

\paragraph{Aligning directions across modalities}
Embedding differences that are parallel to text differences appear in word embeddings~\cite{mikolov2013linguistic}, and directional losses enforce them in CLIP-guided generation and editing~\cite{gal2022stylegannada,kim2022diffusionclip,bar2022text2live} and in PC-CLIP~\cite{sam2025pcclip}. Our pair loss (\cref{sec:method}) is related: it pushes the difference between the video embeddings of a clip and its reversed copy toward the difference between the text embeddings of their two descriptions. Before It Fades~\cite{shin2026fades} also uses a forward-minus-reversed difference, but without training: it injects the difference of hidden states into later layers at inference time. We find that widening the video difference matters more than aligning the two differences, and that the text difference must come from the same pair.

\paragraph{Learning without losing the base}
Continual learning, distillation and backward-compatible training preserve old abilities or embeddings with output, relation or weight constraints~\cite{li2016lwf,kirkpatrick2017ewc,hinton2015distilling,park2019rkd,shen2020bct,wu2022nccl,byun2026adapter}, and frozen teachers protect generalization when adapting CLIP~\cite{khattak2023promptsrc,huang2024froster}.
When fine-tuning CLIP for image classification, keeping the features near the pretrained ones reduces forgetting, by matching the features of several layers (LDIFS~\cite{mukhoti2024ldifs}), distilling on reference data~\cite{zheng2023zscl} or training the head first~\cite{kumar2022lpft}. LDIFS finds the last-layer features alone not enough there. Video embedders change the base only a little, through LoRA~\cite{hu2022lora,halbe2026verve,tv2026vitalx} or a small training set~\cite{bagad2026tara}, but add no output anchor. For MLLM video embedders, we find that holding the output embeddings near the base model, not the weights, keeps general retrieval, even with a loss alone and no reference data. Weight interpolation (WiSE-FT~\cite{wortsman2022wise}) keeps general R@1 through the same closeness (\cref{sec:analysis}).

\section{What Plain Training Changes}
\label{sec:observations}
We study Plain training and then the base models, Qwen3-VL-Embedding-8B (Qwen) and InternVL2-8B-RA (InternVL). Each observation suggests a hypothesis (H1--H3) that becomes one part of our method (\cref{sec:method}).

\paragraph{Benchmarks and notation}
\emph{General retrieval} performance is measured on six benchmarks whose text queries mostly describe what is visible (MSR-VTT, DiDeMo, ActivityNet, VATEX, MSVD, LSMDC). General R@1 is their mean. \emph{Time-sensitive retrieval} performance is measured on RTime, CiA and SSv2-T, whose text queries depend on how actions progress over time.
RTime has $990$ forward test clips and their reversed copies, each with a forward and a reversed text query. The gallery holds the forward clips (RTime Origin) or both (RTime Hard), scored by R@1. In RTime Binary, each query chooses between a clip and its reversed copy (accuracy, chance $50$).
CiA has pairs of opposite actions on SSv2, Charades and EPIC. Each class name is a query, every clip of its class is a positive, and we report mAP. Our main gallery, Chiral, holds the query class and its opposite. Static adds one non-opposite class from every other pair, and All holds every class. CiA-Charades alone means CiA-Charades Chiral.
SSv2-Template (SSv2-T) queries are the $174$ template names (objects hidden). The template names are also training queries (with other clips), so a gain on SSv2-T does not show transfer to unseen text. SSv2-Label uses the same clips with object-specific queries.
We train on SSv2, RTime and MSR-VTT. CiA-Charades, CiA-EPIC and the five other general benchmarks are \emph{unseen}. Qwen has $36$ layers and InternVL $32$. Lk: the output of layer $k$, counted from $0$.

\begin{figure}[t]
  \centering
  \includegraphics[width=.95\linewidth]{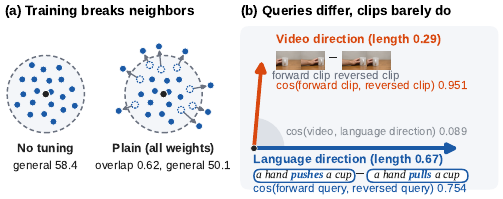}
  \caption{(a) Neighbor overlap of MSR-VTT test clips after Plain training of all weights, with general R@1 (dotted circles: neighbors that left, positions schematic). (b) In the base model, the video direction is less than half as long as the language direction and nearly orthogonal to it (mean of $990$ RTime test pairs, to scale). The pair loss lengthens it (\cref{sec:analysis}).}
  \label{fig:motivation}
\end{figure}

\paragraph{Plain training breaks neighborhoods}
We train with the base losses of \cref{sec:method}, which use reversed and shuffled clips as hard negatives, with no term that ties the model to the base model. We call this Plain training. For each MSR-VTT test clip, we take its $10$ nearest clips by cosine similarity of the output embeddings, before and after training, and call the share that remains the neighbor overlap.
Training all weights of the language model gives a neighbor overlap of only $0.62$ and lowers general R@1 from $58.4$ to $50.1$ (\cref{fig:motivation}a). Low-rank adapters (LoRA) on the middle layers (L12--L23) give $0.65$ and $54.7$.
The clip embeddings also move away from those of the base model (mean cosine $0.81$ with LoRA). This suggests adding a loss that holds the output embeddings near those of the base model (H1).

\paragraph{Language and video differ on direction}
On RTime test pairs, the embeddings of a forward and a reversed query are far apart, while those of a forward and a reversed clip are close (\cref{fig:motivation}b).
We call the forward minus the reversed query embedding the language direction $d_t$, and the forward minus the reversed clip embedding the video direction $d_v$.
Over the $990$ test pairs, the mean norm is $0.669$ for the language directions but only $0.290$ for the video directions, and the two are nearly orthogonal (mean cosine $0.089$). This suggests pushing each clip embedding away from that of its reversed copy (H2).

\begin{figure}[t]
  \centering
  \includegraphics[width=.95\linewidth]{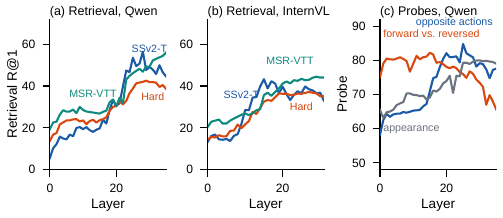}
  \caption{Base models. (a, b) Retrieval from each layer of the frozen model, with a small head per layer trained against the frozen text embeddings (each head has its own training set, so compare curve shapes, not values). (c) Linear probes without text on Qwen: which of two opposite actions a clip shows (CiA-SSv2), which of a clip and its reversed copy is forward (RTime), and, as an appearance control, which action pair a clip shows (CiA).}
  \label{fig:layers_obs}
\end{figure}
\paragraph{Frames are combined in the middle layers}
We measure the performance of the base models layer by layer in three ways (\cref{fig:layers_obs}).
First (\cref{fig:layers_obs}a, b), we train a small retrieval head on each layer of the frozen model: MSR-VTT keeps rising to the output, while SSv2-T and RTime Hard peak near L28 and then fall. InternVL shows a similar but flatter pattern, with SSv2-T peaking earlier (L15).
Second, linear probes without text (\cref{fig:layers_obs}c) check that the peak lies in the video embedding itself. The appearance control stays at $79$ from L25 to the output, while opposite actions peak at L25, and forward versus reversed RTime clips at L15.
Third, we block cross-frame attention at test time: blocking from Lk means that, from layer $k$ on, tokens of one frame cannot attend to those of other frames. On Qwen, blocking from L12 lowers CiA-SSv2 Chiral by $10.6$ and CiA-SSv2 Static by $8.7$, while blocking from L24 changes each score by at most $0.5$. InternVL behaves alike with L8 and L20 (\cref{sec:supp_layers}).
So frames are combined before L24, for appearance and time, and temporal information peaks in or just after the middle layers (L15--L28 on Qwen) and fades toward the output, while appearance does not. This suggests adapting the layers that combine frames, not only the later layers (H3).

\section{Method}
\label{sec:method}
\begin{figure*}[t]
  \centering
  \includegraphics[width=.9\textwidth]{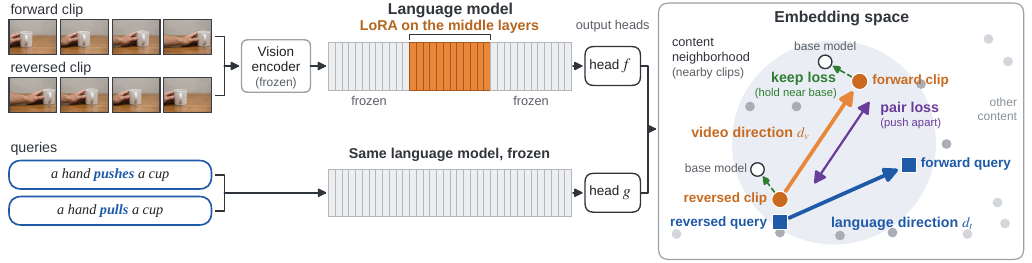}
  \caption{\textbf{Method.} Video passes through the language model with LoRA on the middle layers, text through the same model without LoRA. The output heads $f$ and $g$ give one vector per video clip and per text query. The pair loss makes each query prefer the clip of its own direction. Hollow circles: base-model embeddings of the two clips. Green dashed arrows: the keep loss, which holds each clip near its circle (also applied to text, not drawn). Embedding space: schematic.}
  \label{fig:method}
\end{figure*}
We start from a pretrained embedder (\cref{fig:method}). Two low-rank output heads, $f$ for video and $g$ for text, map each final embedding to the retrieval space: $f(x_v)=\mathrm{norm}(x_v+x_vU_vV_v)$ and $g(x_t)=\mathrm{norm}(x_t+x_tU_tV_t)$, where $\mathrm{norm}$ is L2 normalization, $U_v,U_t\in\mathbb{R}^{d\times r}$, $V_v,V_t\in\mathbb{R}^{r\times d}$, $d$ is the embedding size and $r=256$.\footnote{The factorized form has $2dr$ parameters ($2.1$M for $d=4096$) instead of $d^2$ ($16.8$M), and its update lies in a shared $r$-dimensional subspace, which limits how far the head can move the embeddings.} With $V_v=V_t=0$, both heads start as L2 normalization alone. The text side of the model stays frozen, and each clip and each query is one vector from the final layer.

\paragraph{Base losses} Three losses form the base, as in earlier training with reversed clips~\cite{du2024rtime,bagad2023testoftime}. $\mathcal{L}_\text{nce}$ is a text-to-video InfoNCE loss with temperature $\tau=0.05$, whose negatives include the clip in the opposite direction and a shuffled clip. $\mathcal{L}_\text{template}$ is an InfoNCE loss on SSv2, with template names as queries and clips of the same template as positives. $\mathcal{L}_\text{margin}$, the hard-negative margin loss, is a softplus margin against the same hard negatives, so it pushes each hard negative directly. Training with these three losses alone is called Plain training. From the observations we add two losses and an adapter placement.

\paragraph{Keep loss (H1)} The keep loss holds the video and text embeddings close to those of the base model:
\begin{equation}
  \mathcal{L}_\text{keep}=\bigl[1-\cos(f(x_v),x^0_v)\bigr]+\bigl[1-\cos(g(x_t),x_t)\bigr],
\end{equation}
where $x^0_v$ is the base embedding of the same clip. The text side has no LoRA, so its original embedding is $x_t$. On RTime, the keep loss applies to both the forward and the reversed clip. It anchors the output embeddings, not the weights.

\paragraph{Pair loss (H2)} The pair loss acts on each forward clip and its reversed copy. Only RTime gives a caption for the reversed clip, so the pair loss uses RTime clips. For a forward clip, its reversed copy and their forward and reversed queries, let $x_v,\tilde{x}_v$ be the video embeddings, $x_t,\tilde{x}_t$ the frozen text embeddings, and $s(x_v,x_t)=\cos(f(x_v),g(x_t))$. The forward query should prefer the forward clip, $m=s(x_v,x_t)-s(\tilde{x}_v,x_t)>0$, and the reversed query the reversed clip, $\tilde{m}=s(\tilde{x}_v,\tilde{x}_t)-s(x_v,\tilde{x}_t)>0$. The pair-margin loss pushes both margins:
\begin{equation}
  \mathcal{L}_\text{pm}=\tfrac12\bigl[\varphi(-m)+\varphi(-\tilde{m})\bigr],
\end{equation}
with $\varphi(z)=\mathrm{softplus}(z/\tau)$, the form of $\mathcal{L}_\text{margin}$. It is the part of $\mathcal{L}_\text{margin}$ on each clip and its own reversed copy, given extra weight and averaged over these pairs alone.

The video direction is short and nearly orthogonal to the language direction (\cref{sec:observations}), so a natural variant aligns the two. With the video and language directions $d_v=f(x_v)-f(\tilde{x}_v)$ and $d_t=g(x_t)-g(\tilde{x}_t)$ (\cref{fig:motivation}b, here after the heads), the diff loss
\begin{equation}
  \mathcal{L}_\text{diff}=1-\cos(d_v,d_t)
\end{equation}
turns $d_v$ toward $d_t$. The pair-margin loss does the same implicitly and without normalization: its margins are $m=g(x_t)^\top d_v$ and $\tilde{m}=-g(\tilde{x}_t)^\top d_v$, so $m+\tilde{m}=\langle d_v,d_t\rangle$. We call $\cos(d_v,d_t)$ on test pairs the alignment.

\paragraph{Middle-layer LoRA (H3)} We add LoRA of rank $32$ to layers $12$--$23$ of Qwen and $8$--$19$ of InternVL, where frames are combined. This adds $10.2$M parameters on Qwen. Our objective acts only on the outputs, so we also apply it to all weights of the language model: Ours (all weights).

\paragraph{Objective} We minimize
\begin{equation}\label{eq:objective}
  \mathcal{L}=\underbrace{\mathcal{L}_\text{nce}+w_t\mathcal{L}_\text{template}+\lambda\mathcal{L}_\text{margin}}_\text{base}+\beta\mathcal{L}_\text{keep}+w_p\mathcal{L}_\text{pair},
\end{equation}
where $\mathcal{L}_\text{pair}=\mathcal{L}_\text{pm}$ with $w_p=3$ (Qwen) or $5$ (InternVL), $w_t=1$, $\lambda=1$ and $\beta=10$. This is Ours, with LoRA on the middle layers. Ours (diff loss) uses $\mathcal{L}_\text{diff}$ with $w_p=1$.

\paragraph{Training} We train with AdamW for one epoch on $23{,}955$ clips ($8{,}550$ SSv2, $6{,}555$ RTime, $8{,}850$ MSR-VTT; \cref{sec:supp_setup_data}), batch size $64$. LoRA uses learning rate $10^{-4}$, the output heads $10^{-3}$ and all weights $2\times10^{-5}$.

\section{Experiments}
\label{sec:experiments}
\subsection{Setup}
\cref{sec:observations} defines benchmarks and metrics. CiA follows the official TARA protocol~\cite{bagad2026tara} (test sets in \cref{tab:supp_bench}). All results: text-to-video, one vector per clip, $16$ frames, means over three seeds. Brackets give $95\%$ confidence intervals from a hierarchical bootstrap that resamples seeds and then test items, paired across models (\cref{sec:supp_setup_eval}). With only three seeds, these intervals capture seed-to-seed variation only roughly.

\subsection{Main results}
\begin{table*}[t]
  \caption{Main results (\%, mean$\pm$std of three seeds; columns: \cref{sec:observations}). Bold: best per column and model. $^\ast$Unseen in training. Plain (LoRA): same layers and data, no keep or pair loss. w/o pair loss: $w_p=0$. Ours: base + keep + pair-margin losses ($w_p=3$ on Qwen, $5$ on InternVL), middle-layer LoRA. (all weights): the same loss on all language-model weights. Ours (diff loss): \cref{tab:supp_main_diff}.}
  \label{tab:main}
  \centering\small\setlength{\tabcolsep}{2.6pt}
  \begin{tabular}{@{}lcc|ccc|ccc|cc@{}}
    \toprule
 & \multicolumn{2}{c|}{General} & \multicolumn{3}{c|}{RTime} & \multicolumn{3}{c|}{CiA (Chiral mAP)} & \multicolumn{2}{c}{SSv2} \\
    & Mean of 6 & MSR-VTT & Hard & Binary & Origin & SSv2 & Charades$^\ast$ & EPIC$^\ast$ & Template & Label \\
    \midrule
    \multicolumn{11}{@{}l}{\emph{Qwen3-VL-Embedding-8B}} \\
    No tuning & \textbf{58.4} & 56.1 & 55.5 & 62.8 & \textbf{89.5} & 72.2 & 61.7 & 62.7 & 22.4 & 45.3 \\
    Plain (LoRA) & 54.7{\scriptsize$\pm$0.3} & 53.0{\scriptsize$\pm$0.1} & 57.8{\scriptsize$\pm$0.8} & 68.7{\scriptsize$\pm$0.7} & 84.5{\scriptsize$\pm$0.9} & 87.7{\scriptsize$\pm$0.7} & 71.4{\scriptsize$\pm$0.7} & \textbf{75.6}{\scriptsize$\pm$1.1} & 63.2{\scriptsize$\pm$2.0} & 47.2{\scriptsize$\pm$1.2} \\
    Plain (all weights) & 50.1{\scriptsize$\pm$0.4} & 51.3{\scriptsize$\pm$0.6} & 59.7{\scriptsize$\pm$1.6} & 70.7{\scriptsize$\pm$1.1} & 84.6{\scriptsize$\pm$0.7} & 88.6{\scriptsize$\pm$0.4} & 70.3{\scriptsize$\pm$1.2} & 72.6{\scriptsize$\pm$1.9} & \textbf{64.8}{\scriptsize$\pm$1.2} & 46.3{\scriptsize$\pm$2.2} \\
    Ours w/o pair loss & \textbf{58.4}{\scriptsize$\pm$0.2} & 56.1{\scriptsize$\pm$0.4} & 58.6{\scriptsize$\pm$0.2} & 66.8{\scriptsize$\pm$0.2} & 89.0{\scriptsize$\pm$0.2} & 85.0{\scriptsize$\pm$0.5} & 67.3{\scriptsize$\pm$0.1} & 69.4{\scriptsize$\pm$0.1} & 54.8{\scriptsize$\pm$2.3} & 54.7{\scriptsize$\pm$0.4} \\
    \rowcolor{oursbg}\textbf{Ours} & \textbf{58.4}{\scriptsize$\pm$0.2} & 56.3{\scriptsize$\pm$0.6} & 63.2{\scriptsize$\pm$0.6} & 72.0{\scriptsize$\pm$0.4} & 89.4{\scriptsize$\pm$0.1} & 86.6{\scriptsize$\pm$0.8} & \textbf{72.4}{\scriptsize$\pm$0.2} & 75.3{\scriptsize$\pm$0.3} & 51.9{\scriptsize$\pm$2.3} & 54.2{\scriptsize$\pm$0.4} \\
    \rowcolor{oursbg}\textbf{Ours} (all weights) & 57.5{\scriptsize$\pm$0.3} & \textbf{56.5}{\scriptsize$\pm$0.2} & \textbf{64.2}{\scriptsize$\pm$0.6} & \textbf{73.5}{\scriptsize$\pm$0.6} & 89.4{\scriptsize$\pm$0.3} & \textbf{88.9}{\scriptsize$\pm$0.5} & 72.3{\scriptsize$\pm$0.5} & 75.2{\scriptsize$\pm$1.1} & 64.4{\scriptsize$\pm$2.5} & \textbf{59.1}{\scriptsize$\pm$1.1} \\  %
    \midrule
    \multicolumn{11}{@{}l}{\emph{InternVL2-8B-RA}} \\
    No tuning & 46.0 & 43.9 & 45.3 & 58.0 & 78.4 & 61.1 & 54.8 & 59.1 & 15.5 & 23.8 \\
    Plain (LoRA) & 44.7{\scriptsize$\pm$0.5} & 44.5{\scriptsize$\pm$0.4} & 49.1{\scriptsize$\pm$0.6} & 66.0{\scriptsize$\pm$0.2} & 73.5{\scriptsize$\pm$0.5} & 77.9{\scriptsize$\pm$0.7} & 62.5{\scriptsize$\pm$0.7} & 65.9{\scriptsize$\pm$1.5} & \textbf{53.1}{\scriptsize$\pm$0.3} & 34.2{\scriptsize$\pm$0.8} \\
    Ours w/o pair loss & \textbf{46.8}{\scriptsize$\pm$0.1} & \textbf{45.0}{\scriptsize$\pm$0.2} & 49.2{\scriptsize$\pm$0.9} & 63.2{\scriptsize$\pm$0.4} & \textbf{78.9}{\scriptsize$\pm$0.7} & 74.0{\scriptsize$\pm$0.5} & 59.6{\scriptsize$\pm$0.3} & 64.5{\scriptsize$\pm$0.8} & 46.9{\scriptsize$\pm$2.2} & \textbf{35.7}{\scriptsize$\pm$0.3} \\
    \rowcolor{oursbg}\textbf{Ours} & 46.7{\scriptsize$\pm$0.1} & 44.9{\scriptsize$\pm$0.1} & \textbf{54.0}{\scriptsize$\pm$0.4} & \textbf{69.8}{\scriptsize$\pm$0.2} & 78.4{\scriptsize$\pm$0.2} & \textbf{78.2}{\scriptsize$\pm$0.4} & \textbf{65.3}{\scriptsize$\pm$0.1} & \textbf{68.8}{\scriptsize$\pm$0.4} & 46.2{\scriptsize$\pm$0.9} & 34.0{\scriptsize$\pm$0.4} \\
    \bottomrule
  \end{tabular}
\end{table*}

\begin{figure}[t]
  \centering
  \includegraphics[width=.8\linewidth]{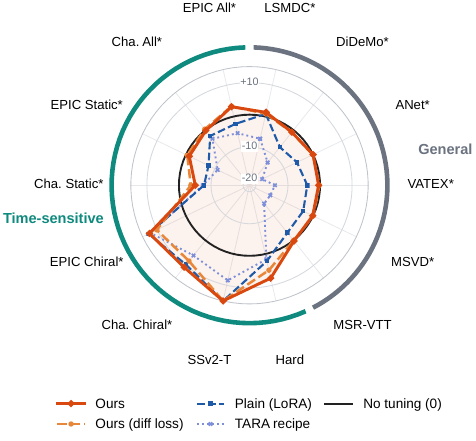}
  \caption{Change from the base model on each benchmark (Qwen, mean of three seeds; R@1, and mAP for CiA). The black circle marks no change: points outside it are gains. Outer arcs group the general and time-sensitive benchmarks. $^*$Unseen in training. Ours and Plain (LoRA) train LoRA on the middle layers. The TARA recipe trains all weights on text only. SSv2-T gains lie beyond the outer ring: Ours (diff loss) $+31.6$, Ours $+29.5$, Plain (LoRA) $+40.8$ (Ours (diff loss): \cref{sec:supp_abl_forms}).}
  \label{fig:radar}
\end{figure}

\cref{tab:main,fig:radar} give the main results (intervals per benchmark: \cref{tab:supp_ci}).
On Qwen, Ours keeps general R@1 and raises time-sensitive retrieval.
Against the base model, the unseen general R@1 (the five general benchmarks other than MSR-VTT) changes by $-0.02$ $[-0.52, 0.48]$, and no single general benchmark changes significantly. Plain (LoRA) loses $3.8$ $[2.9, 4.6]$ on the same five, and Ours is above it on every general benchmark. RTime Hard rises by $+7.7$ $[5.6, 9.9]$ and Binary by $+9.2$ $[7.1, 11.2]$.
The gains also hold on unseen benchmarks: CiA-Charades $+10.7$ $[9.4, 12.0]$ and CiA-EPIC $+12.6$ $[10.7, 14.7]$. On Vinoground~\cite{zhang2024vinoground}, an unseen benchmark whose paired captions differ only in event order, Ours also raises the group score ($+2.4$ $[0.5, 4.3]$). Like the embedding models in~\cite{zhang2024vinoground}, all models stay below chance ($16.7$): a pair counts only if each clip prefers its own caption and each caption its own clip, so a model that prefers one caption for both clips always fails (\cref{sec:supp_vino}).
The CiA gains are on Chiral. Wider galleries give mixed results, and CiA-Charades Static falls by $5.1$ $[3.3, 7.0]$ (all galleries in \cref{sec:supp_cia}).
Ours is slightly above Plain (LoRA) on CiA-Charades Chiral ($+1.0$) and matches it on CiA-EPIC Chiral ($-0.3$); Plain training stays ahead on SSv2-T, whose test queries are the training templates. So on unseen time-sensitive benchmarks Ours gains as much as Plain training but also keeps general retrieval.
On SSv2-Label, whose queries reword the training templates, Ours and Ours w/o pair loss rise by about $9$ points on Qwen while Plain training stays near the base model: the keep loss lets what is learned from the templates transfer to new wording (on InternVL, all trained models rise alike).

About half of the CiA gain comes without the pair loss (\cref{tab:main}).
On top of Ours w/o pair loss, the pair-margin loss adds Hard $+4.6$ $[3.0, 6.2]$, Binary $+5.2$ $[3.7, 6.8]$, CiA-Charades $+5.1$ $[4.4, 6.0]$ and CiA-EPIC $+6.0$ $[4.9, 7.2]$, without lowering general R@1, but not SSv2-T ($-2.9$ $[-8.6, 2.7]$).
InternVL shows the same pattern: unseen general R@1 changes by $+0.7$ $[0.1, 1.2]$ for Ours and by $-1.8$ $[-2.7, -0.8]$ for Plain (LoRA), and time-sensitive retrieval rises.

\paragraph{All weights} Our objective also works on all weights: RTime Hard $64.2$ and general R@1 $57.5$, against $50.1$ for Plain (all weights). LoRA on the middle layers trains $0.15\%$ of the parameters (output heads trained in both) in less than half the training time of all-weight training (\cref{sec:supp_setup_train}). Against all-weight training, it is lower on Hard ($-1.0$ $[-2.9, 0.8]$), equal on unseen CiA and higher on general R@1 ($+0.9$ $[0.3, 1.5]$), so all-weight training goes further along the same trade-off. All weights are also higher on SSv2-T, whose test queries are the training templates.

\subsection{Comparison}
\cref{tab:compare} compares Ours with the TARA recipe, released models and WiSE-FT under our protocol.
The TARA recipe, which fine-tunes on text alone, trades general R@1 for time-sensitive retrieval on the same Qwen model: at learning rate $2\times10^{-5}$ it has general $44.8$ and the best Binary among single models ($75.1$). With the TARA LoRA setting, general R@1 stays at $57.2$ but RTime Hard barely rises ($56.9$), so no TARA setting is above Ours on both general R@1 and RTime Hard.  %
TARA also reports a drop in general retrieval on its own protocol (MSR-VTT NegBench R@5~\cite{alhamoud2025negbench}: $79.8\rightarrow70.2$~\cite{bagad2026tara}).
With our keep loss on the text side, the TARA recipe recovers general R@1 to $57.8$, and Ours is higher on Hard ($+3.4$), CiA-Charades ($+9.9$), CiA-EPIC ($+8.1$) and SSv2-T ($+24.3$).
With a keep loss on both text and video, general reaches $59.2$, the highest in the table, but RTime Hard barely moves ($57.9$), so on general retrieval alone the TARA recipe can do better than Ours.
TARA reaches its best results on Tarsier2 (Tarsier2 alone: general R@1 $12.3$ in our protocol). The released Tarsier2 + TARA is ahead of Ours only on CiA-EPIC, at general R@1 $48.4$, and the other released models are lower on every column (\cref{sec:supp_cmp_tara}).
Ensembles with released TARA ($\oplus$) raise RTime Hard further but need two models and a vector of twice the length. Ours is a better partner than the base model (\cref{sec:supp_cmp_ens}).

\textbf{WiSE-FT.} Interpolating the weights of Plain training and the base model (WiSE-FT~\cite{wortsman2022wise}) also softens the trade-off. With LoRA, Ours lies above the WiSE-FT curve: at equal general R@1, Ours is higher on Hard ($+2.4$ $[1.0, 3.9]$), Binary ($+3.2$ $[1.9, 4.6]$), CiA-Charades ($+3.2$ $[2.3, 4.0]$) and CiA-EPIC ($+3.9$ $[2.6, 5.3]$), while Ours (diff loss) lies on it (\cref{sec:supp_abl_forms}). With all weights, no WiSE-FT point matches Ours (all weights) on both axes (\cref{sec:supp_cmp_wise}). WiSE-FT of Ours w/o keep loss ($\beta=0$) reaches about the same point as Ours (within $1$ point on Hard, Binary and CiA; SSv2-T $-3.3$ $[-9.3, 2.8]$). Both keep the output embeddings near the base model, and general R@1 follows this closeness (\cref{sec:analysis}). The keep loss achieves this in a single run, without choosing $\alpha$.

\subsection{What each part contributes}
\begin{table}[t]
  \caption{Comparison under our protocol (Qwen unless named). General: mean R@1 of six. Hard: RTime Hard. Cha., EPIC: CiA-Charades, CiA-EPIC Chiral. TARA rows: our reruns, with our keep loss (text or both sides) and its LoRA setting. WiSE-FT~\cite{wortsman2022wise}: Plain interpolated with the base model. all w.: all weights. $\oplus$: released Tarsier2 + TARA ensembled with a second model. Bold: best single model; shaded: ours. $^\ast$Unseen in training.}
  \label{tab:compare}
  \centering\footnotesize\setlength{\tabcolsep}{1.8pt}
  \begin{tabular}{@{}lcccccc@{}}
    \toprule
    & General & Hard & Binary & Cha.$^\ast$ & EPIC$^\ast$ & SSv2-T \\
    \midrule
    No tuning & 58.4 & 55.5 & 62.8 & 61.7 & 62.7 & 22.4 \\
    \midrule
    TARA recipe & 44.8 & 57.2 & \textbf{75.1} & 67.8 & 75.0 & 30.8 \\
    \quad + keep loss (text) & 57.8 & 59.8 & 67.3 & 62.5 & 67.2 & 27.6 \\
    \quad + keep loss (text, video) & \textbf{59.2} & 57.9 & 65.5 & 62.0 & 65.7 & 22.8 \\
    TARA (LoRA) & 57.2 & 56.9 & 64.7 & 63.2 & 65.2 & 24.5 \\  %
    \midrule
    CaRe-7B & 46.2 & 48.9 & 59.5 & 53.8 & 61.9 & 23.6 \\
    GVE-7B & 39.9 & 42.8 & 56.1 & 52.2 & 56.9 & 9.8 \\
    LamRA-Ret & 44.8 & 41.9 & 52.8 & 53.0 & 57.9 & 11.5 \\
    Tarsier2 + TARA & 48.4 & 56.5 & 70.8 & 69.0 & \textbf{80.4} & 42.5 \\
    \midrule
    WiSE-FT, LoRA ($\alpha{=}0.5$) & 58.9 & 61.0 & 68.5 & 68.6 & 70.2 & 53.4 \\
    \rowcolor{oursbg}\textbf{Ours} & 58.4 & 63.2 & 72.0 & \textbf{72.4} & 75.3 & 51.9 \\
    WiSE-FT, all w.\ ($\alpha{=}0.5$) & 56.7 & 63.4 & 71.3 & 70.5 & 76.1 & 63.4 \\
    \rowcolor{oursbg}\textbf{Ours} (all w.) & 57.5 & \textbf{64.2} & 73.5 & 72.3 & 75.2 & \textbf{64.4} \\  %
    \midrule
    TARA $\oplus$ Qwen & 58.2 & 63.5 & 71.3 & 67.7 & 76.2 & 36.2 \\
    TARA $\oplus$ Ours & 58.0 & 66.2 & 75.2 & 72.6 & 80.5 & 54.6 \\
    \bottomrule
  \end{tabular}
\end{table}

\label{sec:parts}
\begin{table}[t]
  \caption{Ablations on Qwen3-VL-Embedding-8B (three seeds; all in \cref{sec:supp_abl_full}). Columns as in \cref{tab:compare}. Plain: $\beta=w_p=0$. lr: learning rate. L2-SP: L2 decay toward the initial weights. $\lambda\times5$: five times the hard-negative margin weight. (diff loss): the diff loss ($w_p=1$) as the pair loss. $^\ast$Unseen in training.}
  \label{tab:abl}
  \centering\footnotesize\setlength{\tabcolsep}{1.3pt}
  \begin{tabular}{@{}lcccccc@{}}
    \toprule
    & General & Hard & Binary & Cha.$^\ast$ & EPIC$^\ast$ & SSv2-T \\
    \midrule
    \rowcolor{oursbg}\textbf{Ours} ($w_p{=}3$, $\beta{=}10$) & 58.4 & 63.2 & 72.0 & 72.4 & 75.3 & 51.9 \\
    \midrule
    \multicolumn{7}{@{}l}{\emph{Keeping general retrieval}} \\
    Plain & 54.7 & 57.8 & 68.7 & 71.4 & 75.6 & 63.2 \\
    Plain + L2-SP & 56.6 & 56.4 & 65.6 & 67.3 & 68.0 & 45.4 \\
    $\beta{=}0$, lr$\times\frac15$ (diff loss) & 56.1 & 57.6 & 67.9 & 69.4 & 72.6 & 53.6 \\
    $\beta{=}3$, no pair & 58.1 & 59.2 & 68.0 & 69.9 & 73.4 & 63.0 \\
    $\beta{=}3$ (diff loss) & 57.7 & 60.7 & 70.5 & 72.9 & 76.8 & 61.9 \\
    \midrule
    \multicolumn{7}{@{}l}{\emph{Pair loss}} \\
    Ours w/o pair loss & 58.4 & 58.6 & 66.8 & 67.3 & 69.4 & 54.8 \\
    w/o pair loss, $\lambda{\times}5$ & 57.9 & 60.0 & 69.2 & 71.2 & 73.4 & 53.1 \\
    Diff loss, $w_p{=}5$ & 57.8 & 62.1 & 72.3 & 73.4 & 76.7 & 52.3 \\
    Pair margin, $w_p{=}5$ & 58.5 & 64.1 & 73.0 & 73.3 & 76.4 & 49.2 \\
    \bottomrule
  \end{tabular}
\end{table}

\begin{figure}[t]
  \centering
  \includegraphics[width=.72\linewidth]{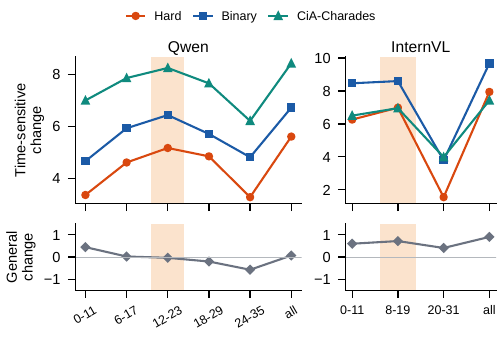}
  \caption{LoRA placement with Ours (diff loss), as change from the base model (three seeds; pair-margin loss: see text). Late-only LoRA (L24--L35 on Qwen, L20--L31 on InternVL) gains less, and all layers gain about as much as the middle layers. Shaded: ours.}
  \label{fig:placement}
\end{figure}
\paragraph{Keeping general retrieval} $\beta$ in \cref{eq:objective} moves the model along a frontier (\cref{tab:supp_abl_full,fig:supp_frontier}): with the diff loss, general R@1 rises from $53.8$ at $\beta=0$ to $58.6$ at $\beta=30$, while SSv2-T falls from $62.8$ to $50.6$. Training without the keep loss at one fifth of the learning rate, which moves the weights less, reaches general R@1 $56.1$, while the keep loss with $\beta=3$ at the full learning rate reaches $57.7$ (both with the diff loss). Without the pair loss, L2-SP~\cite{li2018l2sp} gives $56.6$ (Plain: $54.7$; keep loss at $\beta=3$: $58.1$) and is lower than the keep loss on every time-sensitive column. L2-SP and the lower learning rate move the LoRA weights $3$--$6\times$ less than Plain training, while the keep loss moves them about as much (Frobenius norm $2.02$ vs.\ $2.14$) but in another direction (cosine $0.17$ with Plain training, vs.\ $0.29$ between its seeds; \cref{sec:supp_keep_dw}), and of these, only the keep loss preserves general R@1. So general retrieval depends on where the output embeddings go, not on how far the weights move. We hold out no validation split, so $\beta$, $w_p$ and the WiSE-FT $\alpha$ were set on test. Two checks suggest that this does not drive our conclusions. First, a rule fixed in advance and applied to $450$ clips held out from training also selects our $\beta$ and $w_p$ (by its tie-break, as it barely separates the settings). Second, every tried $\beta\geq3$ and $w_p$ keeps general R@1 within $0.7$ of the base model and is above it on RTime Hard and unseen CiA (\cref{tab:supp_heldout}). WiSE-FT is compared over its whole $\alpha$ curve (\cref{sec:supp_cmp_wise}).

\paragraph{Without general-domain data} Removing MSR-VTT, which leaves all six general benchmarks unseen, lowers general R@1 only to $57.4$ (base model $58.4$), while Plain (LoRA) falls to $49.7$. Ours w/o pair loss reaches $57.1$, so the keep loss does this. The pair loss adds time-sensitive retrieval at the same general R@1 (Hard $+4.4$, CiA $+4.8$ / $+5.5$; all in \cref{sec:supp_abl_full}).

\paragraph{The pair loss} The pair loss raises RTime and unseen CiA retrieval at the same general R@1 (\cref{tab:abl}), and five times the hard-negative margin weight ($\lambda\times5$) gains less at a similar general R@1. The language direction must be the pair's own: giving each clip that of another pair removes the gain (diff loss; \cref{sec:supp_abl_diff}). Without normalization, the pair-margin loss also lengthens the video direction $d_v$ more than the diff loss ($\lVert d_v\rVert$ $0.48$ at $w_p=5$, from $0.29$ in the base model), and RTime Hard follows this length more than the alignment: at $w_p=5$ the diff loss has the higher alignment but lower Hard ($62.1$ vs.\ $64.1$; \cref{tab:dir}). With a criterion fixed before the runs, the pair-margin loss matches the diff loss at the same weight on unseen CiA, on both models (\cref{sec:supp_abl_forms}).

\paragraph{Placement} Consistent with H3, adapting only the late layers (\cref{fig:placement}), which come after frames are combined and where temporal information fades, gains less: with the diff loss, late-only LoRA reaches RTime Hard $58.8$ against $60.7$ for the middle layers on Qwen, and $46.8$ against $52.2$ on InternVL. With the pair-margin loss, the middle layers are also above late-only LoRA on general R@1 and unseen CiA, and match all layers with a third of the layers (\cref{sec:supp_why_place}). In Ours, blocking cross-frame attention from L12 keeps only $44\%$ and $32\%$ of the gain on CiA-Charades and CiA-EPIC, while blocking from L24 keeps it, so the gain relies on the adapted layers. InternVL behaves alike, and for Ours (all weights), most of the gain survives blocking from L12 (\cref{sec:supp_why_block}).  %

\section{Why It Works}
\label{sec:analysis}
\begin{figure}[t]
  \centering
  \includegraphics[width=.9\linewidth]{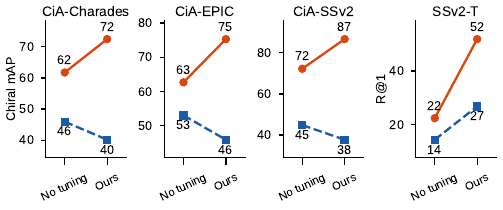}
  \caption{Reversal test, Ours on Qwen (three seeds; Chiral mAP for CiA, R@1 for SSv2-T). Solid: original test clips. Dashed: the same clips reversed, scored against the original text. A reversed clip often shows the opposite action, so a lower dashed line means more use of frame order. On CiA, Ours raises the solid line and lowers the dashed one. On SSv2-T, both rise.}
  \label{fig:reversal}
\end{figure}
\paragraph{Reversal} We reverse the test clips, keep the text queries and labels, and evaluate again (\cref{fig:reversal}). A model that uses frame order should then score lower, and one that ignores order should score the same. We write the gains over the base model as $\Delta_\text{orig}=a+b$ and $\Delta_\text{rev}=a-b$, where $a$ is independent of order and $b$ flips sign under reversal. The order share $b/\Delta_\text{orig}=(\Delta_\text{orig}-\Delta_\text{rev})/(2\Delta_\text{orig})$ is $0$ when reversal leaves the gain unchanged and $1$ when it turns the gain into an equal loss. Ours has $77\%$ and $78\%$ on CiA-Charades and CiA-EPIC (about $80\%$ on InternVL) but $29\%$ on SSv2-T, whose gain is from the template loss (\cref{sec:supp_abl_parts}). Variants with and without the pair loss depend on order alike ($76$--$85\%$, \cref{sec:supp_why_reversal}), so the pair loss adds magnitude, not a different kind of gain. Shuffling the frames brings Ours down to the base model on CiA, so this gain depends on coherent frame order (other frame changes: \cref{sec:supp_why_frames}).

\paragraph{Where general retrieval is lost} We compare $28$ variants of Qwen (keep-loss weights $\beta$, pair losses, learning rates, L2-SP, all weights and WiSE-FT; $82$ models). On five general benchmarks, we measure how close the clip embeddings stay to those of the base model (mean cosine) and how many of their $10$ nearest neighbors remain. General R@1 follows the closeness ($r=0.92$) more than the neighbor overlap ($r=0.81$) or the weight change, and given the overlap or the weight change, only closeness keeps a positive partial correlation. Closeness and overlap are themselves correlated ($r=0.93$), so this is a tendency consistent with H1, not a separation (\cref{sec:supp_keep_neigh,sec:supp_keep_dw}). WiSE-FT, which adds no loss on the outputs, follows the same relation. The clips of Ours stay close to their base embeddings (mean cosine $0.984$, against $0.811$ for Plain (LoRA)) and keep most of their $10$ nearest neighbors ($0.89$). With the embeddings held this close, the pair loss mainly widens the gap between the video embeddings of a forward clip and its reversed copy: closeness is the same without it, while the video direction (\cref{fig:motivation}b) grows from $0.29$ in the base model to $0.34$ without the pair loss and $0.45$ with it (\cref{tab:dir,sec:supp_why_agree}).

\section{Conclusion and Limitations}
\label{sec:conclusion}
With one vector per clip, training for time-sensitive retrieval can keep general R@1. General R@1 tracks how close the output embeddings stay to those of the base model, not how far the weights move. This holds whether a keep loss keeps them close during training or weight interpolation restores them after training. Keeping the weights near their initial values during training (L2-SP or a lower learning rate) preserves less general R@1 than the keep loss. With a keep loss, a pair-margin loss that pushes each clip away from its reversed copy, and LoRA on the middle layers, a strong embedder gains about $8$ to $13$ points on RTime Hard and on opposite actions in unseen CiA, keeps most of its general retrieval even without general-domain data, and lies above WiSE-FT of Plain training. Middle-layer LoRA ($0.15\%$ of the parameters) nearly matches training all weights on RTime and unseen CiA and keeps general retrieval better.

\paragraph{Limitations} The trade-off is softened, not removed: Plain (LoRA) and all-weight training are higher on some time-sensitive columns, and $\beta$ moves along a frontier. The SSv2-T gain comes from training on the test template names, so it does not show generalization. On unseen CiA, the gain is on opposite actions (Chiral): All moves little, and Static can fall, mainly on classes with the same verb and another object (on these classes by $10.4$ and $7.4$ points against the base model on CiA-Charades and CiA-EPIC, \cref{sec:supp_cia}). Some analyses have one seed, the middle-layer peak is less distinct on InternVL, and our protocol differs from those of other papers.

\paragraph{Use of Generative AI} We used Claude Code (Anthropic) for coding, analysis and writing. The authors take responsibility for all content.

{\small\bibliographystyle{ieeenat_fullname}\bibliography{refs}}
\backrefsetup{disable}   %
\makeatletter\setlength{\@fptop}{0pt}\makeatother   %
\clearpage
\makeatletter\immediate\write\@auxout{\string\gdef\string\mainlastpage{\number\numexpr\value{page}-1\relax}}\makeatother
\appendix
\noindent{\Large\bfseries Appendix\par}\vspace{0.8em}

\section{Setup Details}
\label{sec:supp_setup}
This section collects the settings behind all results of the main text (\cref{sec:method,sec:experiments}). Differences are computed from unrounded values.

\subsection{Training settings}
\label{sec:supp_setup_train}
\paragraph{Model and adapters} We add LoRA of rank $32$ to layers $12$--$23$ of Qwen and $8$--$19$ of InternVL, which adds $10.2$M trainable parameters on Qwen ($4.2$M more for the two output heads). The language model gets no LoRA on the text side, so text embeddings come from the frozen model; only the text head $g$ is trained. Two low-rank output heads, $f$ for video and $g$ for text, map each final embedding to the retrieval space, with rank $256$ (\cref{sec:method}). Ours (all weights) applies the loss of Ours to all weights of the language model instead of LoRA: the $36$ decoder layers ($6.9$B parameters; the token embeddings and the final norm stay frozen), so LoRA trains $0.15\%$ of them; text embeddings again come from the frozen model. The TARA recipe also trains the token embeddings ($7.6$B parameters). %
LoRA scales its update by $1$ (\texttt{lora\_alpha} equal to the rank, $32$) and uses no dropout, with $A$ initialized as in Kaiming-uniform and $B=0$. LoRA adapts the query, key, value and output projections of self-attention, not the MLP; on InternVL the fused $qkv$ projection gets three separate adapters. The base model runs in bf16 with gradient checkpointing, while LoRA and the heads keep fp32 weights. Each head is $f(x)=\mathrm{norm}(x+xUV)$ with $U\sim\mathcal{N}(0,1/d)$ and $V=0$, so both heads start as L2 normalization alone. The embedding is the last-token state after the final norm, L2-normalized. Qwen uses the instruction ``Retrieve videos relevant to the user's query.'' for both queries and clips, and keeps the pixel budget of the embedder (about $1{,}300$ tokens per clip). InternVL resizes each frame to $448\times448$ ($256$ tokens) and uses its prompts ``Summary above video in one word:'' and ``Summary above sentence in one word:''.

\paragraph{Optimization} We train with AdamW for one epoch, with batch size $64$. LoRA uses learning rate $10^{-4}$, the output heads $10^{-3}$ and all-weight training $2\times10^{-5}$. The weights of the objective are $w_t=1$, $\lambda=1$ and $\beta=10$, with $w_p=3$ on Qwen and $5$ on InternVL for the pair-margin loss and $w_p=1$ for the diff loss; the InfoNCE temperature is $\tau=0.05$.
The learning rate follows a cosine decay from its peak to $0.1$ of it over one epoch ($374$ steps), multiplied by a linear warm-up over the first $50$ steps. Weight decay is $0$ for LoRA and all-weight training and $0.01$ for the heads, and gradients are clipped to norm $1$. The hard-negative margin loss $\mathcal{L}_\text{margin}$ is $\mathrm{softplus}((s_\text{neg}-s_\text{pos})/\tau)$ with no margin constant. Each RTime and SSv2 query has two hard negatives: the reversed copy of its clip and one frame-shuffled copy of the forward clip (for the query of a reversed RTime clip: the forward clip and the shuffled copy). MSR-VTT clips enter in forward order only, and their queries have no hard negative. A shuffle is a random non-identity permutation of the $16$ frames, drawn anew for each clip and epoch. The template loss uses the SSv2 clips of each batch, with all clips of a template as positives. A batch of $64$ source videos holds about $146$ clips (with the reversed and shuffled copies) and $82$ queries. %

\paragraph{Compute} On one A100 ($80$\,GB), Ours on Qwen trains in $5.5$ hours with gradient checkpointing from L12 (peak GPU memory $64$\,GiB), and one evaluation on all benchmarks takes $1.9$--$2.3$ hours. Ours (all weights) does not fit in GPU memory with the setup of Ours: it checkpoints every layer, uses gradient caching (one extra forward pass per step) and keeps the fp32 master weights and both AdamW moments of its $6.9$B trained parameters on the CPU ($77.6$\,GiB); it trains in $14.4$ hours on the same GPU type (peak GPU memory $38$\,GiB). Part of the gap in training time between Ours and Ours (all weights) comes from these memory-saving settings. InternVL Ours, with gradient caching, takes $25$ to $28$ hours. %

\paragraph{Longer training} One epoch is the default for every model. Training for three epochs (one seed; Ours on Qwen, Ours (diff loss) on InternVL) raises RTime Hard and Binary, CiA-Charades and SSv2-T further on both models, and CiA-EPIC only on InternVL, while RTime Origin falls slightly. General R@1 stays within $0.7$ of the one-epoch run of the same seed ($57.9$ vs.\ $58.6$ on Qwen, $47.1$ vs.\ $46.8$ on InternVL), so the keep loss holds general retrieval over longer training. %

\paragraph{Choices made on test} We hold out no validation split for these choices. $\beta=10$ came from early small-scale runs on the same benchmarks; \cref{tab:supp_abl_full} shows $\beta\in\{0,3,10,30\}$. For the pair-margin loss we tried $w_p\in\{3,5\}$ on Qwen and use $3$ (\cref{tab:abl} reports both values); on InternVL we use $5$. For L2-SP we ran strengths $30$ and $300$ and report $300$, whose general R@1 on one seed was closer to Ours ($56.9$ vs.\ $56.5$). For WiSE-FT, \cref{tab:compare} shows $\alpha=0.5$, the point with the highest RTime Hard on both curves; $\alpha=0.75$ is higher on some other time-sensitive columns but lower on general R@1 (\cref{tab:supp_wise}). The Qwen layer band (L12--23) was fixed before the placement sweep. On InternVL, the band L8--19 comes from an earlier half-epoch comparison with L20--31 (two seeds); it covers the layers where blocking cross-frame attention lowers time-sensitive retrieval (\cref{sec:observations,sec:supp_layers}). %

\paragraph{Selection on held-out clips} %
To check the choice of $\beta$ and $w_p$ made on test, we fixed a rule before the pair-margin $\beta=3$ runs started and applied it to the $450$ clips held out from training (\cref{sec:supp_setup_data}; $150$ each from MSR-VTT, RTime with both directions, and SSv2; final-embedding R@1 against all held-out clips). A setting is eligible if its held-out MSR-VTT R@1 is at most $1.0$ below the base model. Among eligible settings, the rule picks the highest mean of RTime and SSv2 R@1; if the best two differ by less than $0.5$, it picks the larger $\beta$, i.e., the stronger keep loss. The check is not blind: when writing the rule, we knew the held-out numbers of the $\beta=10$ runs and an earlier diff-loss comparison in which $\beta=3$ was ahead on these clips. All four settings with $\beta\in\{3,10\}$ and $w_p\in\{3,5\}$ are eligible, and their scores span only $0.6$ (\cref{tab:supp_heldout}): $\beta=3$, $w_p=3$ scores highest ($76.7$) and Ours ($\beta=10$, $w_p=3$) second ($76.4$), so the tie-break picks Ours. On test, all four keep general R@1 within $0.7$ of the base model and are above it on RTime Hard and CiA Chiral, but below it on CiA Static ($21.6$--$24.7$ vs.\ $29.8$ on CiA-Charades and $28.1$--$32.1$ vs.\ $33.3$ on CiA-EPIC; not in the table).  %

\begin{table*}[t]
  \centering
  \caption{Selection of $\beta$ and $w_p$ on the $450$ held-out clips (Qwen, pair-margin loss, LoRA on the middle layers, mean over three seeds). Held-out: R@1 per source; score: mean of RTime and SSv2. Test columns as in \cref{tab:compare}. The rule picks Ours (shaded).}
  \label{tab:supp_heldout}
  \small
  \begin{tabular}{@{}lcccc|cccccc@{}}
    \toprule
    & \multicolumn{4}{c|}{Held-out} & \multicolumn{6}{c}{Test} \\
    & MSR-VTT & RTime & SSv2 & Score & General & Hard & Binary & Cha.$^\ast$ & EPIC$^\ast$ & SSv2-T \\
    \midrule
    No tuning & 74.0 & 61.3 & 74.7 & 68.0 & 58.4 & 55.5 & 62.8 & 61.7 & 62.7 & 22.4 \\
    \midrule
    \rowcolor{oursbg}Ours ($\beta=10$, $w_p=3$) & 74.9 & 71.6 & 81.3 & 76.4 & 58.4 & 63.2 & 72.0 & 72.4 & 75.3 & 51.9 \\
    $\beta=10$, $w_p=5$ & 74.4 & 72.4 & 80.0 & 76.2 & 58.5 & 64.1 & 73.0 & 73.3 & 76.4 & 49.2 \\
    $\beta=3$, $w_p=3$ & 75.3 & 72.1 & 81.3 & 76.7 & 57.8 & 63.1 & 72.9 & 74.4 & 78.2 & 58.2 \\
    $\beta=3$, $w_p=5$ & 76.0 & 71.1 & 81.1 & 76.1 & 57.8 & 63.3 & 72.9 & 74.8 & 78.9 & 53.8 \\
    \bottomrule
  \end{tabular}
\end{table*}  %

\subsection{Data}
\label{sec:supp_setup_data}
We train on $23{,}955$ clips: $8{,}550$ from SSv2 (a $50$-per-template subset of the training split), $6{,}555$ RTime training clips that have a reversed caption, and $8{,}850$ MSR-VTT 9k training videos with $5$ of their $20$ captions. From each source, $150$ clips are held out to monitor training and for the selection check of \cref{tab:supp_heldout}, never for training; $60$ RTime training clips could not be downloaded. Runs that leave one source out train one epoch over the remaining clips ($236$ to $271$ steps instead of $374$). No test clip is used for training. On RTime, each training clip enters with its reversed copy and its forward and reversed queries; on SSv2, template names are the queries of the template loss. \Cref{tab:supp_bench} lists the test sets. %
\begin{table}[h]
  \caption{Test sets. RTime: $990$ of the $1{,}000$ test videos (the others could not be downloaded), each with its reversed copy. SSv2-T: $174$ templates (objects hidden) with $12$ clips each. DiDeMo and ActivityNet: the sentences of a video are joined into one paragraph; DiDeMo uses the first $31$ seconds; one of its $1{,}004$ test videos is unreadable and is left out for all models. SSv2-Label: the SSv2-T clips with object-specific queries. VATEX: the $1{,}303$ available clips of the HGR test split. Same domain: the training split of the dataset is used for training, its test clips are not.}
  \label{tab:supp_bench}
  \centering\footnotesize\setlength{\tabcolsep}{3pt}
  \begin{tabular}{@{}lrrll@{}}
    \toprule
    Benchmark & Clips & Queries & Metric & Training \\
    \midrule
    RTime & $990$ & $1{,}980$ & R@1 / acc. & same domain \\
    SSv2-T & $2{,}088$ & $174$ & R@1 & same domain \\
    SSv2-Label & $2{,}088$ & $1{,}989$ & R@1 & same domain \\
    CiA-SSv2 & $1{,}430$ & $32$ & mAP & same domain \\
    CiA-Charades & $5{,}498$ & $56$ & mAP & unseen \\
    CiA-EPIC & $3{,}108$ & $132$ & mAP & unseen \\
    MSR-VTT 1k-A & $1{,}000$ & $1{,}000$ & R@1 & same domain \\
    DiDeMo & $1{,}003$ & $1{,}003$ & R@1 & unseen \\
    ActivityNet & $4{,}917$ & $4{,}917$ & R@1 & unseen \\
    VATEX & $1{,}303$ & $13{,}030$ & R@1 & unseen \\
    MSVD & $670$ & $27{,}763$ & R@1 & unseen \\
    LSMDC & $1{,}000$ & $1{,}000$ & R@1 & unseen \\
    \bottomrule
  \end{tabular}
\end{table}

\subsection{Evaluation protocol}
\label{sec:supp_setup_eval}
All retrieval results after \cref{sec:observations} are text-to-video (Vinoground also scores video-to-text, \cref{sec:supp_vino}), with one vector per clip and per query, from the final layer. All models read $16$ frames. On RTime, the reversed clip is the same $16$ frames in reverse order. RTime Origin and Hard rank a gallery (R@1); Binary is a choice between a forward clip and its reversed copy (accuracy, chance $50$). SSv2-T and the general benchmarks report R@1, and general R@1 is the mean over the six general benchmarks.  %
The $16$ frames are sampled uniformly over the clip, or over the annotated segment, at indices $\mathrm{round}(\mathrm{linspace}(0,n-1,16))$, the same rule as in training. Test queries and clips use the training instructions.

\paragraph{CiA protocol} CiA follows the official TARA protocol~\cite{bagad2026tara}: each class name is a query, every clip of the query class is a positive, and we report AP per class query, averaged over the queries. Chiral: the gallery holds the query class and its opposite class. Static: the gallery holds the query class and one non-opposite class from every other pair. All: the gallery holds every class. CiA is thus class-level retrieval, close to zero-shot classification. For example, Static ranks ``putting a blanket somewhere'' against ``putting a broom somewhere''.
Our one change to the TARA code: our CiA-SSv2 queries fill the object noun of each template, while the raw TARA templates make several classes share one query string. Chiral is unaffected.

\paragraph{Confidence intervals} Brackets give $95\%$ confidence intervals from a hierarchical bootstrap ($10{,}000$ resamples) in two stages: first the three seeds of each model, then the test units, paired across models. The units are clips for the general benchmarks (a clip can have several queries), queries for RTime, SSv2-T and SSv2-Label, and classes for CiA. We report the mean over three seeds unless marked.

\paragraph{Benchmarks one by one} Against the base model, no single general benchmark changes significantly for Ours on Qwen (from DiDeMo, $-0.8$ $[-2.4, 0.7]$, to LSMDC, $+1.3$ $[-0.3, 2.9]$), and Ours is above Plain (LoRA) on every general benchmark (\cref{tab:supp_ci}). On InternVL, Ours is above both the base model and Plain (LoRA) on general R@1, RTime Hard and Binary and the Chiral columns of both unseen CiA benchmarks (CiA-Charades, CiA-EPIC), equal to the base model on RTime Origin, and below Plain (LoRA) on SSv2-T.

\paragraph{Charades} As an extra benchmark outside general R@1, we evaluate Charades text-to-video retrieval on Qwen (test split, $1{,}863$ videos, the script of each video as its query, $16$ frames over the whole video; final checkpoints of the three seeds, as in the main tables). No model trains on Charades. Ours keeps Charades R@1 ($30.8\pm0.1$ against $29.8$ for the base model), as do Ours w/o pair loss ($30.6\pm0.1$) and TARA (LoRA) ($30.3\pm0.1$). Ours (all weights) has $28.9\pm0.3$, Plain (LoRA) $26.1\pm0.5$, the TARA recipe $23.9\pm1.4$ and Plain (all weights) $16.3\pm1.1$, so Charades broadly follows general R@1. Its test videos include all source videos of CiA-Charades, so the two benchmarks are not independent.  %

\begin{table*}[t]
  \caption{Per-benchmark results of Ours (mean over three seeds) with $95\%$ hierarchical bootstrap intervals. Bold: the interval excludes $0$ (before rounding).} %
  \label{tab:supp_ci}
  \centering\footnotesize
  \emph{Qwen3-VL-Embedding-8B}\\[2pt]
  \begin{tabular}{@{}lccc|cc@{}}
\toprule
 & Ours & No tuning & Ours $-$ No tuning & Plain (LoRA) & Ours $-$ Plain (LoRA) \\
\midrule
MSR-VTT & 56.3 & 56.1 & $+0.2$ $[-1.2, 1.6]$ & 53.0 & \textbf{\boldmath$+3.3$ $[1.4, 5.3]$} \\
DiDeMo & 58.8 & 59.6 & $-0.8$ $[-2.4, 0.7]$ & 52.8 & \textbf{\boldmath$+6.0$ $[4.0, 8.1]$} \\
ActivityNet & 60.9 & 60.9 & $+0.0$ $[-0.7, 0.7]$ & 55.3 & \textbf{\boldmath$+5.6$ $[4.3, 6.8]$} \\
VATEX & 75.6 & 76.1 & $-0.5$ $[-1.0, 0.1]$ & 72.1 & \textbf{\boldmath$+3.5$ $[2.7, 4.5]$} \\ %
MSVD & 63.1 & 63.3 & $-0.1$ $[-0.8, 0.5]$ & 59.8 & \textbf{\boldmath$+3.3$ $[2.2, 4.5]$} \\
LSMDC & 35.5 & 34.2 & $+1.3$ $[-0.3, 2.9]$ & 35.0 & $+0.5$ $[-1.1, 2.1]$ \\
Mean of 6 & 58.4 & 58.4 & $+0.0$ $[-0.5, 0.5]$ & 54.7 & \textbf{\boldmath$+3.7$ $[3.0, 4.4]$} \\
\midrule
RTime Hard & 63.2 & 55.5 & \textbf{\boldmath$+7.7$ $[5.6, 9.9]$} & 57.8 & \textbf{\boldmath$+5.4$ $[3.7, 7.2]$} \\
RTime Binary & 72.0 & 62.8 & \textbf{\boldmath$+9.2$ $[7.1, 11.2]$} & 68.7 & \textbf{\boldmath$+3.2$ $[1.6, 4.9]$} \\
RTime Origin & 89.4 & 89.5 & $-0.1$ $[-1.4, 1.2]$ & 84.5 & \textbf{\boldmath$+4.9$ $[3.3, 6.6]$} \\
\midrule
CiA-Charades Chiral & 72.4 & 61.7 & \textbf{\boldmath$+10.7$ $[9.4, 12.0]$} & 71.4 & \textbf{\boldmath$+1.0$ $[0.0, 2.0]$} \\
CiA-Charades Static & 24.7 & 29.8 & \textbf{\boldmath$-5.1$ $[-7.0, -3.3]$} & 22.0 & \textbf{\boldmath$+2.7$ $[1.8, 3.7]$} \\
CiA-Charades All & 19.6 & 19.8 & $-0.2$ $[-1.6, 1.2]$ & 17.5 & \textbf{\boldmath$+2.2$ $[1.4, 3.0]$} \\
CiA-EPIC Chiral & 75.3 & 62.7 & \textbf{\boldmath$+12.6$ $[10.7, 14.7]$} & 75.6 & $-0.3$ $[-2.1, 1.2]$ \\
CiA-EPIC Static & 32.1 & 33.3 & $-1.1$ $[-3.4, 1.1]$ & 25.4 & \textbf{\boldmath$+6.7$ $[4.4, 9.2]$} \\
CiA-EPIC All & 26.2 & 23.1 & \textbf{\boldmath$+3.1$ $[1.3, 5.0]$} & 20.7 & \textbf{\boldmath$+5.5$ $[3.5, 7.6]$} \\
\midrule
SSv2-T & 51.9 & 22.4 & \textbf{\boldmath$+29.5$ $[21.8, 37.2]$} & 63.2 & \textbf{\boldmath$-11.3$ $[-18.4, -4.2]$} \\
SSv2-Label & 54.2 & 45.3 & \textbf{\boldmath$+8.9$ $[7.2, 10.7]$} & 47.2 & \textbf{\boldmath$+7.1$ $[5.0, 9.1]$} \\
\bottomrule
\end{tabular}
\\[6pt]
  \emph{InternVL2-8B-RA}\\[2pt]
  \begin{tabular}{@{}lccc|cc@{}}
\toprule
 & Ours & No tuning & Ours $-$ No tuning & Plain (LoRA) & Ours $-$ Plain (LoRA) \\
\midrule
MSR-VTT & 44.9 & 43.9 & $+1.0$ $[-0.5, 2.4]$ & 44.5 & $+0.3$ $[-1.5, 2.2]$ \\
DiDeMo & 44.3 & 43.2 & $+1.2$ $[-0.6, 3.0]$ & 41.2 & \textbf{\boldmath$+3.1$ $[0.9, 5.4]$} \\
ActivityNet & 52.4 & 52.2 & $+0.2$ $[-0.6, 1.0]$ & 48.2 & \textbf{\boldmath$+4.2$ $[2.8, 5.5]$} \\
VATEX & 63.0 & 62.7 & $+0.3$ $[-0.5, 1.0]$ & 59.3 & \textbf{\boldmath$+3.6$ $[2.6, 4.6]$} \\
MSVD & 51.6 & 50.9 & $+0.7$ $[-0.3, 1.8]$ & 50.7 & $+1.0$ $[-0.4, 2.3]$ \\
LSMDC & 24.3 & 23.4 & $+0.9$ $[-0.7, 2.4]$ & 24.1 & $+0.2$ $[-1.5, 1.9]$ \\
Mean of 6 & 46.7 & 46.0 & \textbf{\boldmath$+0.7$ $[0.2, 1.2]$} & 44.7 & \textbf{\boldmath$+2.1$ $[1.3, 2.8]$} \\
\midrule
RTime Hard & 54.0 & 45.3 & \textbf{\boldmath$+8.7$ $[6.5, 11.0]$} & 49.1 & \textbf{\boldmath$+4.8$ $[3.1, 6.6]$} \\
RTime Binary & 69.8 & 58.0 & \textbf{\boldmath$+11.8$ $[9.4, 14.1]$} & 66.0 & \textbf{\boldmath$+3.8$ $[2.3, 5.4]$} \\
RTime Origin & 78.4 & 78.4 & $+0.0$ $[-1.9, 1.9]$ & 73.5 & \textbf{\boldmath$+4.9$ $[2.9, 7.0]$} \\
\midrule
CiA-Charades Chiral & 65.3 & 54.8 & \textbf{\boldmath$+10.6$ $[9.0, 12.2]$} & 62.5 & \textbf{\boldmath$+2.8$ $[1.7, 3.9]$} \\
CiA-Charades Static & 25.6 & 23.0 & \textbf{\boldmath$+2.6$ $[1.2, 4.2]$} & 28.2 & \textbf{\boldmath$-2.6$ $[-3.5, -1.7]$} \\
CiA-Charades All & 17.8 & 12.7 & \textbf{\boldmath$+5.1$ $[4.0, 6.2]$} & 19.0 & \textbf{\boldmath$-1.2$ $[-1.8, -0.6]$} \\
CiA-EPIC Chiral & 68.8 & 59.1 & \textbf{\boldmath$+9.7$ $[7.9, 11.7]$} & 65.9 & \textbf{\boldmath$+2.9$ $[1.0, 5.0]$} \\
CiA-EPIC Static & 18.6 & 19.1 & $-0.5$ $[-2.3, 1.2]$ & 19.2 & $-0.7$ $[-2.4, 0.9]$ \\
CiA-EPIC All & 13.5 & 11.9 & \textbf{\boldmath$+1.5$ $[0.4, 2.7]$} & 14.0 & $-0.6$ $[-1.8, 0.7]$ \\
\midrule
SSv2-T & 46.2 & 15.5 & \textbf{\boldmath$+30.7$ $[23.0, 38.3]$} & 53.1 & \textbf{\boldmath$-6.9$ $[-13.0, -0.8]$} \\
SSv2-Label & 34.0 & 23.8 & \textbf{\boldmath$+10.2$ $[8.5, 11.9]$} & 34.2 & $-0.2$ $[-1.9, 1.6]$ \\
\bottomrule
\end{tabular}

\end{table*}

\subsection{TARA rebuild}  %
\label{sec:supp_setup_tara}
\paragraph{TARA recipe on Qwen} The $20$K training triplets of TARA are not released, so we rebuild them from license-free sources with the mix of the paper ($20{,}000$ text triplets: $8{,}000$ core, $1{,}000$ negation, $1{,}000$ time, $10{,}000$ multimodal). Core triplets come from the SimCSE NLI set, as in TARA; negation triplets are NLI triplets whose hard negative adds a negation, selected by a rule instead of an LLM; time triplets come from RTime training captions instead of Ego4D narrations; multimodal triplets come from WebVid-CoVR with the videos replaced by their captions. We copy the released trainer: contrastive loss with temperature $0.05$ over groups of $256$ triplets, optimizer batch $768$, AdamW with learning rate $2\times10^{-5}$, $10\%$ warm-up and linear decay, two epochs ($52$ steps), on all weights of the language model. The text format is that of our protocol, since the prompt TARA used on Qwen3-VL-Embedding is not released. Since its time triplets come from RTime training captions, the rebuild sees in-domain text for RTime, which favors the TARA recipe, not Ours, on RTime. To measure this effect, we also train the TARA recipe once with the $1{,}000$ time triplets replaced by $1{,}000$ further core triplets (one seed; an ablation, not a baseline). Against the full recipe with the same seed, it is higher on general R@1 ($47.3$ vs.\ $44.5$) and RTime Origin ($83.0$ vs.\ $80.3$) and lower on RTime Hard ($53.8$ vs.\ $56.5$), Binary ($66.3$ vs.\ $74.7$), CiA-Charades ($61.8$ vs.\ $68.1$), CiA-EPIC ($67.6$ vs.\ $74.9$) and SSv2-T ($28.2$ vs.\ $30.5$). Each value lies outside the range of the full recipe over three seeds. Without time triplets, CiA-Charades falls to the level of the base model ($61.8$ vs.\ $61.7$), RTime Hard falls below it ($53.8$ vs.\ $55.5$) and general R@1 stays far below it ($47.3$ vs.\ $58.4$), so the time triplets carry most of the gains of the TARA recipe on RTime and CiA, but not its loss on general retrieval.  %
All TARA rows use the same rebuilt set (one build) and the two-epoch TARA schedule, with three seeds. Lower learning rates for TARA trace the same trade-off between general and time-sensitive retrieval ($5\times10^{-6}$: general $51.1$, RTime Hard $60.4$; $2\times10^{-6}$: $53.7$, $57.4$; three seeds each). The keep variants of TARA add our keep loss with $\beta=10$ on the text side, or on text and video; the video term uses $64$ clips of our training set per step. For TARA (LoRA), the TARA paper gives no rank, target layers or learning rate, so we use rank $32$ on the attention and MLP projections of all $36$ layers, on both the video and the text side ($87.3$M parameters), and learning rate $2\times10^{-5}$, chosen to favor TARA; a first run with $2\times10^{-4}$ (one seed) gave general R@1 $50.4$ and RTime Hard $58.8$. %

\section{Full Ablation Table and Pair-Loss Controls}
\label{sec:supp_abl}
This section supports \cref{sec:parts}: it lists the ablations on Qwen and gives further controls of the pair loss and of the parts of the loss.

\subsection{Full ablation table}
\label{sec:supp_abl_full}
\Cref{tab:supp_abl_full} gives the ablations on Qwen; \cref{tab:abl} is a subset. The pair-margin settings with $\beta=3$ are in \cref{tab:supp_heldout}, and the pair-margin model without the keep loss ($\beta=0$) and its WiSE-FT are in \cref{tab:supp_dw} and \cref{sec:supp_cmp_wise}.
Removing SSv2 lowers SSv2-T to $25.9$ but not RTime Hard or unseen CiA. Removing RTime lowers RTime Hard to $58.7$; this run uses the diff loss on SSv2 pairs, as the pair-margin loss needs RTime.
\begin{table*}[t]
  \caption{Ablations on Qwen (mean over three seeds; $^\S$one seed). Pair margin: the pair-margin loss ($\mathcal{L}_\text{pm}$, \cref{sec:method}). Inner product: the diff loss with $\langle d_v,d_t\rangle$ instead of the cosine. $\lambda{\times}k$: $k$ times the weight $\lambda$ of the hard-negative margin loss. Plain: no keep loss, no pair loss. Unless the row name says otherwise, rows under Parts of the loss, Target of the diff loss and Keep weight are variants of Ours (diff loss). General: mean R@1 over the six general benchmarks. Hard and Binary: RTime. Cha., EPIC: CiA-Charades and CiA-EPIC Chiral. $^\ast$Unseen in training. $^\dagger$The pair loss needs RTime, so the row without RTime uses SSv2 pairs of opposite actions instead.}
  \label{tab:supp_abl_full}
  \centering\footnotesize\setlength{\tabcolsep}{6pt}
  \begin{tabular}{@{}lcccccc@{}}
    \toprule
    & General & Hard & Binary & Cha.$^\ast$ & EPIC$^\ast$ & SSv2-T \\
    \midrule
    \rowcolor{oursbg}\textbf{Ours} (pair margin, $w_p{=}3$, $\beta{=}10$) & 58.4 & 63.2 & 72.0 & 72.4 & 75.3 & 51.9 \\
    Ours (diff loss, $w_p{=}1$) & 58.3 & 60.7 & 69.3 & 70.0 & 72.8 & 54.0 \\
    \midrule
    \multicolumn{7}{@{}l}{\emph{Forms and weights of the pair loss}} \\
    Ours w/o pair loss ($w_p{=}0$) & 58.4 & 58.6 & 66.8 & 67.3 & 69.4 & 54.8 \\
    Diff loss, $w_p{=}5$ & 57.8 & 62.1 & 72.3 & 73.4 & 76.7 & 52.3 \\
    w/o pair loss, $\lambda{\times}5$ & 57.9 & 60.0 & 69.2 & 71.2 & 73.4 & 53.1 \\
    w/o pair loss, $\lambda{\times}2$ & 58.2 & 59.3 & 67.7 & 68.6 & 71.0 & 53.1 \\
    w/o pair loss, $\lambda{\times}10$ & 57.2 & 60.5 & 70.8 & 72.9 & 75.1 & 51.1 \\
    Inner product, weight 1$^\S$ & 58.5 & 59.3 & 67.8 & 68.6 & 71.5 & 53.4 \\
    Inner product, weight 5$^\S$ & 56.5 & 56.0 & 71.3 & 73.6 & 77.4 & 49.4 \\
    Inner product, weight 20 & 32.4 & 28.6 & 70.8 & 72.6 & 83.4 & 26.2 \\
    Pair margin, $w_p{=}5$ & 58.5 & 64.1 & 73.0 & 73.3 & 76.4 & 49.2 \\
    Pair margin + diff loss & 58.2 & 63.4 & 72.5 & 73.1 & 76.4 & 51.3 \\
    \midrule
    \multicolumn{7}{@{}l}{\emph{Parts of the loss}} \\
    No margin & 58.6 & 60.0 & 68.2 & 68.7 & 71.5 & 54.6 \\
    No template loss & 58.1 & 61.9 & 70.9 & 71.0 & 72.6 & 25.7 \\
    Reversed neg.\ only$^\S$ & 58.7 & 62.0 & 70.1 & 70.7 & 74.1 & 52.9 \\
    No hard neg.\ or template & 58.9 & 63.0 & 71.1 & 68.1 & 69.8 & 28.9 \\
    Fixed text head & 58.6 & 58.6 & 66.6 & 67.6 & 70.0 & 49.4 \\
    LoRA, no heads & 58.6 & 58.5 & 66.5 & 66.6 & 69.1 & 42.1 \\
    Heads, no LoRA & 57.2 & 56.7 & 65.2 & 66.2 & 67.6 & 43.3 \\
    \midrule
    \multicolumn{7}{@{}l}{\emph{Target of the diff loss}} \\
    Shuffled target & 58.4 & 58.5 & 66.6 & 67.4 & 69.6 & 52.9 \\
    Diff loss, $w_p{=}5$, shuffled rev.\ query & 58.0 & 60.0 & 69.2 & 70.2 & 72.3 & 54.0 \\
    Generated rev.\ queries & 58.3 & 60.7 & 69.4 & 71.5 & 75.5 & 54.6 \\
    Generated, shuffled rev.\ query & 58.2 & 60.7 & 69.4 & 70.6 & 73.9 & 53.6 \\
    Generated, diff loss, $w_p{=}5$ & 55.9 & 60.6 & 72.0 & 75.3 & 82.0 & 49.8 \\
    Generated, pair margin, $w_p{=}5$ & 57.7 & 63.8 & 73.0 & 74.7 & 78.7 & 52.3 \\
    \midrule
    \multicolumn{7}{@{}l}{\emph{Keep weight}} \\
    $\beta=0$ & 53.8 & 58.1 & 70.7 & 73.9 & 78.2 & 62.8 \\
    $\beta=0$, $w_p=0$ (Plain) & 54.7 & 57.8 & 68.7 & 71.4 & 75.6 & 63.2 \\
    $\beta=3$ & 57.7 & 60.7 & 70.5 & 72.9 & 76.8 & 61.9 \\
    $\beta=3$, $w_p=0$ & 58.1 & 59.2 & 68.0 & 69.9 & 73.4 & 63.0 \\
    $\beta=30$ & 58.6 & 58.5 & 66.3 & 66.3 & 68.1 & 50.6 \\
    $\beta=0$, learning rate $\times\frac15$ & 56.1 & 57.6 & 67.9 & 69.4 & 72.6 & 53.6 \\
    Plain + L2-SP & 56.6 & 56.4 & 65.6 & 67.3 & 68.0 & 45.4 \\
    Neighbor keep loss & 58.2 & 59.0 & 67.3 & 70.0 & 70.4 & 60.3 \\
    \midrule
    \multicolumn{7}{@{}l}{\emph{All weights}} \\
    \rowcolor{oursbg}\textbf{Ours} (all weights) & 57.5 & 64.2 & 73.5 & 72.3 & 75.2 & 64.4 \\
    Ours (diff loss, all weights) & 57.9 & 63.3 & 72.1 & 71.2 & 73.9 & 65.7 \\
    Plain (all weights) & 50.1 & 59.7 & 70.7 & 70.3 & 72.6 & 64.8 \\
    \midrule
    \multicolumn{7}{@{}l}{\emph{Training data (one set removed)}} \\
    Without SSv2 & 58.6 & 63.0 & 71.8 & 72.6 & 76.6 & 25.9 \\
    Without MSR-VTT & 57.4 & 63.0 & 71.4 & 72.5 & 75.0 & 53.6 \\
    Without SSv2 (diff loss) & 58.6 & 61.3 & 70.2 & 71.0 & 74.4 & 24.1 \\
    Without RTime (diff loss)$^\dagger$ & 58.5 & 58.7 & 66.6 & 68.0 & 70.3 & 55.0 \\
    Without MSR-VTT (diff loss) & 57.0 & 60.0 & 68.5 & 69.9 & 72.1 & 55.4 \\
    Without MSR-VTT, w/o pair loss & 57.1 & 58.6 & 66.4 & 67.7 & 69.6 & 54.0 \\
    Without MSR-VTT, Plain & 49.7 & 58.1 & 68.8 & 71.6 & 75.6 & 61.7 \\
    \bottomrule
  \end{tabular}
\end{table*}

\subsection{The pair loss: further controls}
\label{sec:supp_abl_diff}
These controls complete the paragraph ``The pair loss'' of \cref{sec:parts}.

\paragraph{Diff loss against stronger hard negatives} At matched general R@1 ($57.8$ and $57.9$), the diff loss at $w_p=5$ is above $\lambda\times5$ on Hard ($+2.1$ $[0.3, 3.8]$), Binary ($+3.1$ $[1.3, 4.9]$), CiA-Charades ($+2.2$ $[1.3, 3.0]$) and CiA-EPIC ($+3.3$ $[1.5, 5.1]$). %
$\lambda\times10$ lowers general R@1 to $57.2$ and stays below the diff loss at $w_p=5$ on Hard, Binary and CiA-EPIC, while $\lambda\times10$ nearly reaches the diff loss at $w_p=5$ on CiA-Charades ($72.9$ vs.\ $73.4$).
At $w_p=1$ the evidence is weaker: Ours (diff loss) matches $\lambda\times5$ on Hard and Binary, is lower on CiA, and keeps $0.5$ more general R@1.

\paragraph{Target of the pair loss} In the shuffled-target control, each clip gets the language direction of another RTime pair (diff loss, $w_p=1$). The shuffled-target control gives RTime Hard $58.5$ and CiA-Charades / CiA-EPIC Chiral $67.4$ / $69.6$, the level of Ours w/o pair loss ($58.6$, $67.3$ / $69.4$), so the pair loss must act on the clip's own pair.
When only the reversed query comes from another pair, the diff loss at $w_p=5$ falls to about the level of $\lambda\times5$ (three seeds; row ``shuffled rev.\ query'' in \cref{tab:supp_abl_full}).  %

\paragraph{Form of the diff loss} With an inner product instead of the cosine, general R@1 falls as the weight grows ($58.5$, $56.5$ and $32.4$ at weights $1$, $5$ and $20$; one seed at weights $1$ and $5$, three seeds at weight $20$), so the diff loss needs the normalization of the cosine.

\paragraph{Balance of Chiral and Static} On wider CiA galleries, the diff loss at $w_p=5$ also loses retrieval among other actions (CiA-Charades Static $21.5$ against $25.3$ for $\lambda\times5$; CiA-EPIC Static $27.4$ against $32.1$), so the diff loss at $w_p=5$ shifts the balance toward opposite actions rather than raising every column.
Ours (pair-margin loss, $w_p=3$) lies between Ours (diff loss) and the diff loss at $w_p=5$ in its balance of Chiral and Static, and has a higher RTime Hard ($63.2$) than Ours (diff loss) and the diff loss at $w_p=5$.

\paragraph{Same weight on InternVL} With $w_p=5$ for both forms of the pair loss (pair margin and diff loss; three seeds each), the pair-margin loss matches the diff loss on InternVL, as on Qwen, under the criterion fixed before the runs (\cref{sec:supp_abl_forms}): CiA-Charades / CiA-EPIC Chiral $65.3$ / $68.8$ against $65.1$ / $69.6$ for the diff loss (within $1.0$), Static $25.6$ / $18.6$ against $23.8$ / $17.2$ and All $17.8$ / $13.5$ against $16.4$ / $13.0$ (not lower). The pair-margin loss is also higher on general R@1 ($46.7$ vs.\ $46.2$), RTime Hard ($54.0$ vs.\ $51.5$) and SSv2-T ($46.2$ vs.\ $44.4$). \Cref{tab:cia_full_ivl} gives all galleries.

\subsection{Two forms of the pair loss}
\label{sec:supp_abl_forms}
For the two forms we fixed a criterion before the runs: the pair-margin loss matches the diff loss at the same weight if both unseen CiA Chiral columns are within $1.0$ and CiA Static and All are not lower. With $w_p=5$ the pair-margin loss meets it on both models (InternVL: \cref{sec:supp_abl_diff}), and is also higher on RTime Hard and general R@1; on SSv2-T it is lower on Qwen and higher on InternVL. At their default weights (Ours vs.\ Ours (diff loss)), at the same general R@1, the pair-margin loss favors RTime Hard and unseen CiA Chiral, and the diff loss CiA Static and SSv2-T (\cref{fig:radar}). Adding the diff loss to the pair-margin loss keeps general R@1 and RTime and raises unseen CiA Chiral only slightly ($+0.7$ $[0.2, 1.1]$ and $+1.1$ $[0.3, 2.1]$).  %
\Cref{tab:supp_main_diff} gives the rows of Ours (diff loss) that \cref{tab:main} leaves out. Ours (diff loss) lies on the WiSE-FT curve of Plain (LoRA) (\cref{sec:supp_cmp_wise}).  %
\begin{table*}[t]
  \caption{Rows of Ours (diff loss) left out of \cref{tab:main} (mean$\pm$std over three seeds; same columns). The diff loss ($w_p=1$) replaces the pair-margin loss as the pair loss. $^\ast$Unseen in training.}
  \label{tab:supp_main_diff}
  \centering\small\setlength{\tabcolsep}{2.6pt}
  \begin{tabular}{@{}lcc|ccc|ccc|cc@{}}
    \toprule
 & \multicolumn{2}{c|}{General} & \multicolumn{3}{c|}{RTime} & \multicolumn{3}{c|}{CiA (Chiral mAP)} & \multicolumn{2}{c}{SSv2} \\
    & Mean of 6 & MSR-VTT & Hard & Binary & Origin & SSv2 & Charades$^\ast$ & EPIC$^\ast$ & Template & Label \\
    \midrule
    \multicolumn{11}{@{}l}{\emph{Qwen3-VL-Embedding-8B}} \\
    Ours (diff loss) & 58.3{\scriptsize$\pm$0.3} & 56.1{\scriptsize$\pm$0.4} & 60.7{\scriptsize$\pm$0.3} & 69.3{\scriptsize$\pm$0.4} & 89.0{\scriptsize$\pm$0.1} & 86.1{\scriptsize$\pm$0.2} & 70.0{\scriptsize$\pm$0.1} & 72.8{\scriptsize$\pm$0.3} & 54.0{\scriptsize$\pm$2.1} & 54.7{\scriptsize$\pm$0.4} \\
    Ours (diff loss, all weights) & 57.9{\scriptsize$\pm$0.2} & 56.3{\scriptsize$\pm$0.2} & 63.3{\scriptsize$\pm$0.6} & 72.1{\scriptsize$\pm$0.8} & 88.9{\scriptsize$\pm$0.2} & 88.4{\scriptsize$\pm$0.3} & 71.2{\scriptsize$\pm$1.5} & 73.9{\scriptsize$\pm$0.9} & 65.7{\scriptsize$\pm$2.9} & 58.5{\scriptsize$\pm$0.7} \\
    \midrule
    \multicolumn{11}{@{}l}{\emph{InternVL2-8B-RA}} \\
    Ours (diff loss) & 46.8{\scriptsize$\pm$0.2} & 44.9{\scriptsize$\pm$0.2} & 52.2{\scriptsize$\pm$1.0} & 66.6{\scriptsize$\pm$0.9} & 78.9{\scriptsize$\pm$0.6} & 76.4{\scriptsize$\pm$0.7} & 61.7{\scriptsize$\pm$0.1} & 66.4{\scriptsize$\pm$0.9} & 46.6{\scriptsize$\pm$3.0} & 35.2{\scriptsize$\pm$0.4} \\
    \bottomrule
  \end{tabular}
\end{table*}

\subsection{Reversed clips alone as negatives}
\label{sec:supp_abl_revneg}
The InfoNCE negatives include the reversed and the shuffled copies of each query's clip. Reversed clips alone as negatives give a slightly higher result (RTime Hard $62.0$ against $60.6$ for Ours (diff loss) with the same seed, and higher on every time-sensitive column; one seed; row ``Reversed neg.\ only'' in \cref{tab:supp_abl_full}), so the shuffled copies add nothing measurable here. %

\subsection{Parts of the loss}
\label{sec:supp_abl_parts}
Removing one part at a time from Ours (diff loss) (rows under Parts of the loss in \cref{tab:supp_abl_full}) shows that the hard negatives help transfer to unseen CiA, the SSv2-T gain comes from the template loss ($54.0\to25.7$ without it), and the drop on CiA Static comes mainly from the output heads.

\paragraph{Output heads} With the text output head $g$ fixed at the identity (three seeds), general R@1 stays at $58.6$, but against Ours (diff loss) Hard falls by $2.1$, Binary by $2.6$, CiA Chiral by $2.4$ and $2.8$ (CiA-Charades, CiA-EPIC) and SSv2-T by $4.6$, while CiA-Charades and CiA-EPIC Static rise by $2.9$ and $3.7$. The trained text head carries part of the time-sensitive gain of Ours (diff loss) and part of its Static drop.
Against the base model, LoRA without heads raises CiA Chiral ($66.6$, $69.1$) and changes CiA Static by $+0.2$ and $+3.0$; heads without LoRA gain less on Chiral ($66.2$, $67.6$) and lower Static by $6.3$ and $5.2$ (CiA-Charades, CiA-EPIC).

\section{Generated Reversed Queries}
\label{sec:supp_gen}
This section gives a further control of the pair loss (paragraph ``The pair loss'' in \cref{sec:parts}): whether reversed queries generated for SSv2 and MSR-VTT clips give the pair loss a role for language.

\paragraph{Generation} We generate reversed queries with Qwen3-VL-8B-Instruct from the training caption alone (text only, greedy decoding). The prompt asks for a caption of the same clip played backwards: reverse the direction of every action, swap the order of events, keep the people, objects and setting, do not mention the reversal, and answer ``NONE'' when reversal would not change the description (a static scene, talking, a symmetric motion). The prompt gives seven hand-written examples, none from RTime. For SSv2 clips whose template has an opposite, the reversed query is the opposite template filled with the clip's own objects, without the LLM. Outputs equal to the caption count as NONE. About $87\%$ of the SSv2 captions and half of the MSR-VTT captions get a reversed query. Examples: ``uncovering pencil'' $\to$ ``covering pencil''; ``a video of two women hugging'' $\to$ ``two women releasing their hug''. On $100$ RTime training captions with an annotated reversed caption, the generated query retrieves its annotated counterpart among the $100$ at R@1 for $98$ of the $99$ captions with an answer (one answer is NONE). %

\paragraph{Results} In Ours (diff loss, $w_p=1$) the generated queries enter only the diff loss, so the margin losses are unchanged. Against Ours (diff loss) without generated queries, general R@1 stays at $58.3$ and RTime Hard and Binary do not change, while CiA-Charades rises by $+1.5$ $[1.1, 1.9]$ and CiA-EPIC by $+2.7$ $[1.9, 3.6]$. When the generated reversed query comes from another clip, the CiA-Charades and CiA-EPIC gains are $+0.6$ and $+1.1$ (three seeds). %
At $w_p=5$ we compare the diff loss and the pair-margin loss (three seeds each). With the diff loss, generated queries change CiA Chiral by $+2.0$ / $+5.3$ (CiA-Charades / CiA-EPIC), general R@1 by $-1.8$, RTime Hard by $-1.5$ and CiA Static by $-2.3$ / $-2.8$. With the pair-margin loss, also applied to the generated pairs, generated queries change these values by $+1.4$ / $+2.3$, $-0.8$, $-0.3$ and $-1.4$ / $-1.4$. The diff loss thus gains more than the pair-margin loss on CiA-EPIC Chiral ($+3.1$) but little on CiA-Charades ($+0.6$), so it meets the criterion we had fixed before the runs (at least $1.0$ more on both benchmarks) only on CiA-EPIC.  %

\section{Keep Loss Analysis}
\label{sec:supp_keep}
This section supports the keep loss (H1): the paragraph ``Keeping general retrieval'' of \cref{sec:parts} and the analysis of where general retrieval is lost in \cref{sec:analysis}.

\subsection{Frontier of the keep weight}
\label{sec:supp_keep_frontier}
\begin{figure}[h]
  \centering
  \includegraphics[width=\linewidth]{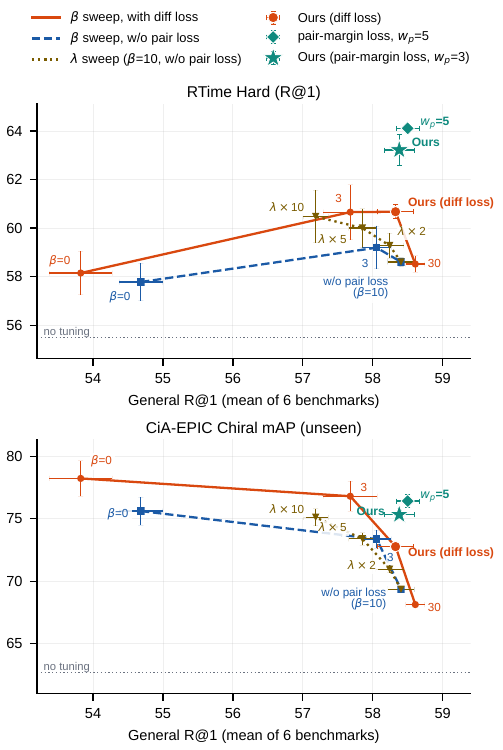}  %
  \caption{General R@1 against RTime Hard (top) and CiA-EPIC Chiral (bottom) on Qwen (mean $\pm$ std over three seeds). Solid and dashed lines sweep the keep weight $\beta$ with the diff loss and without the pair loss, respectively; the dotted line sweeps the hard-negative margin weight $\lambda$ at $\beta=10$ without the pair loss. Star: Ours (pair-margin loss, $w_p=3$); diamond: $w_p=5$. Gray dotted horizontal line: No tuning.}
  \label{fig:supp_frontier}
\end{figure}

\Cref{fig:supp_frontier} shows the frontier on Qwen. With the diff loss, raising $\beta$ from $0$ to $30$ moves general R@1 from $53.8$ to $58.6$, while RTime Hard is highest at $\beta=3$ and $10$ ($60.7$). The two pair-margin points (Ours and $w_p=5$) lie above all three sweeps at their general R@1. On RTime Hard they are above every swept point ($63.2$ and $64.1$); on CiA-EPIC Chiral, some swept points score higher, but only at lower general R@1 ($53.8$ to $57.7$).

\subsection{Neighborhoods and closeness}
\label{sec:supp_keep_neigh}
\begin{figure}[h]
  \centering
  \includegraphics[width=\linewidth]{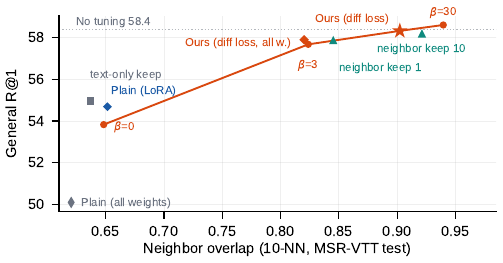}
  \caption{Neighbor overlap against general R@1 on Qwen. Neighbor overlap is the share of the $10$ nearest clips of each MSR-VTT test clip that remain from the base model. The line sweeps the keep weight $\beta$. Neighbor keep 1, 10: weight of the neighbor keep loss. Text-only keep: the keep loss on the text side only. The neighbor keep loss lies on the same curve, and Plain (all weights) lies below this curve. The correlations in the text use the mean over five general benchmarks.}
  \label{fig:supp_neighbours}
\end{figure}
\paragraph{Closeness or neighbors} General R@1 follows the neighbor overlap (\cref{fig:supp_neighbours}), and general R@1 follows closeness to the base embeddings even more. We measure both on five general benchmarks (DiDeMo is left out: the saved DiDeMo embeddings of the base model and of the trained runs do not cover the same clips). Over $28$ variants of Qwen (keep weights, pair losses, learning rates, L2-SP, all-weight training and WiSE-FT mixtures of Plain training, of Ours (diff loss) and of the pair-margin model without the keep loss; up to three seeds each, $82$ models), $r$ is $0.92$ for the mean cosine to the base embedding and $0.81$ for the $10$-nearest-neighbor overlap; the two measures correlate at $0.93$. Over the $28$ variant means, the partial correlation is $0.79$ for closeness given the overlap and $-0.32$ for the overlap given closeness, so closeness leads; since the two measures are themselves correlated, this is a tendency consistent with H1, not a separation. WiSE-FT of models trained without the keep loss lies on or above the line fitted on closeness (residuals $0.0$ to $+1.6$ R@1): interpolation keeps general R@1 as its closeness predicts, or slightly more. Measured instead on held-out training clips (not the general test clips), closeness still follows general R@1 ($r=0.92$ $[0.81, 0.98]$ over the means of $17$ variants, $49$ models with a held-out pass), so the relation between closeness and general R@1 is not an artifact of measuring on the test clips.  %

Among runs with a keep loss, closeness and neighbor overlap can hardly be told apart (correlation with general R@1 $r=0.95$ and $0.94$). Training with one fifth of the learning rate (diff loss, $\beta=0$) shows that the two measures can differ: this run has about the overlap of Plain (LoRA), $0.65$ vs.\ $0.64$, but stays closer to the base model ($0.85$ vs.\ $0.81$) and keeps more general R@1 ($56.1$ vs.\ $54.7$). A forward and a reversed clip stay close (cosine $0.95$ in the base model, $0.92$ in Ours (diff loss), $0.85$ without the keep loss (Plain (LoRA))), so the video direction between a clip and its reversed copy grows as a local move. The higher neighbor overlap with the keep loss also holds beyond MSR-VTT: on the other general benchmarks and on InternVL, the neighbor overlap is $0.58$--$0.71$ for Plain (LoRA) and $0.86$--$0.92$ for Ours (diff loss; one seed), although individual benchmarks can still rise without the keep loss, such as LSMDC. %

\paragraph{Neighbor keep loss} We replace the video keep loss with a neighbor keep loss, which keeps only the original neighborhood among the clips of a batch (diff loss, $w_p=1$). Without any video keep loss, general R@1 is $55.0$, neighbor overlap $0.63$ and closeness $0.79$. The neighbor keep loss restores these three values to $58.2$, $0.92$ and $0.97$. A constraint on neighbors alone also keeps general R@1, but our runs cannot separate neighbors from closeness, because closeness returns with the neighbors. At a tenth of the neighbor keep weight, the three values are $57.9$, $0.84$ and $0.91$ (one seed). For comparison, $\beta=3$ keeps closeness better ($0.96$) but fewer neighbors ($0.82$), with general $57.7$. %

\paragraph{Limits} Neighbors are not the whole story: Plain (LoRA) and Plain (all weights) have about the same neighbor overlap ($0.64$ and $0.61$ over five general benchmarks; $0.65$ and $0.62$ on MSR-VTT, as in \cref{sec:observations}), yet their general R@1 differs ($54.7$, $50.1$), and so does their closeness ($0.81$, $0.67$).

\subsection{Weight change}
\label{sec:supp_keep_dw}
We measure how far the backbone weights move as the Frobenius norm of the whole update, $\lVert\Delta W\rVert_F$, taken jointly over all adapted matrices (for LoRA, $BA$; for all weights, every matrix of the language model; for WiSE-FT, $\alpha$ times the update of its source run). \Cref{tab:supp_dw} gives the values. Over the $82$ models, general R@1 correlates with closeness to the base embeddings ($r=0.92$) far more than with $\lVert\Delta W\rVert_F$ ($r=-0.47$); the partial correlation is $0.91$ $[0.79, 0.95]$ for closeness given the weight change and $-0.22$ $[-0.50, 0.15]$ for the weight change given closeness (intervals from resampling the $28$ variants), and in a joint regression over the variant means closeness explains general R@1 (standardized coefficients $+0.89$ and $-0.08$ for $\log\lVert\Delta W\rVert_F$). The six all-weight variants alone cannot separate closeness from the weight change, since $\alpha$ moves both together. %

The norm ignores the direction of the update, so we also compare directions: the cosine between two updates $\Delta W$ (Frobenius inner product over the $48$ adapted matrices) and the overlap of their column spaces (mean $\cos^2$ of the top-$8$ principal angles). Runs with the same seed share the initialization of the LoRA matrices $A$, so we compare runs across seeds. Two seeds of Plain (LoRA) have cosine $0.29$; Ours, Ours (diff loss) and Ours w/o pair loss have $0.17$, $0.18$ and $0.18$ to the update of Plain training, below their own seed-to-seed values ($0.30$--$0.33$), while the diff loss without the keep loss ($\beta=0$) has $0.28$, the level of Plain training itself. The column spaces of Ours and Plain training still overlap ($0.44$, against $0.56$ between seeds of Plain training and $0.02$--$0.08$ for random subspaces). So the keep loss turns the weight update to another direction of about the same size; L2-SP finds its own direction, which repeats across seeds ($0.90$). %
\begin{table}[h]
  \caption{Weight change, closeness to the base embeddings (mean cosine) and general R@1 (mean over three seeds). $\beta$ rows and the learning-rate row use the diff loss, except the two rows of Ours w/o keep loss (pair-margin loss). All w.: all weights.}
  \label{tab:supp_dw}
  \centering\footnotesize\setlength{\tabcolsep}{3pt}
  \begin{tabular}{@{}lccc@{}}
    \toprule
    & $\lVert\Delta W\rVert_F$ & Closeness & General \\
    \midrule
    Plain (LoRA) & 2.14 & 0.811 & 54.7 \\
    Ours & 2.02 & 0.984 & 58.4 \\
    Ours (diff loss) & 2.09 & 0.986 & 58.3 \\
    Ours w/o pair loss & 2.10 & 0.987 & 58.4 \\
    $\beta=0$ & 2.17 & 0.813 & 53.8 \\
    $\beta=30$ & 2.06 & 0.995 & 58.6 \\
    Plain + L2-SP & 0.37 & 0.902 & 56.6 \\
    Learning rate $\times\frac15$ & 0.75 & 0.850 & 56.1 \\
    WiSE-FT, LoRA ($\alpha=0.5$) & 1.07 & 0.940 & 58.9 \\
    Ours w/o keep loss ($\beta=0$) & 2.12 & 0.809 & 54.7 \\
    \quad + WiSE-FT ($\alpha=0.5$) & 1.06 & 0.937 & 58.9 \\
    Plain (all weights) & 16.5 & 0.673 & 50.1 \\
    Ours (all weights) & 15.8 & 0.941 & 57.5 \\
    Ours (diff loss, all weights) & 15.5 & 0.949 & 57.9 \\
    WiSE-FT, all w.\ ($\alpha=0.5$) & 8.3 & 0.817 & 56.7 \\
    \bottomrule
  \end{tabular}
\end{table}

\subsection{Weight anchors: learning rate and L2-SP}
\label{sec:supp_keep_lr}
Two controls keep the weights closer to their initial values instead of the output embeddings (three seeds each; rows in \cref{tab:supp_abl_full}). With the diff loss and no keep loss ($\beta=0$), one fifth of the learning rate (for LoRA and the heads) gives general $56.1$ and Hard $57.6$, against $53.8$ / $58.1$ at the full rate and $57.7$ / $60.7$ with the keep loss at $\beta=3$. Without the pair loss, an L2 anchor of the trained weights (Plain + L2-SP~\cite{li2018l2sp}, implemented as decoupled weight decay on the LoRA matrices and the head, which pulls the update $BA$ and the head update toward zero) gives general $56.6$ and Hard $56.4$, against $54.7$ / $57.8$ for Plain (LoRA) and $58.1$ / $59.2$ for the keep loss at $\beta=3$. We tried strengths $30$ and $300$ on one seed and kept the strength with general R@1 closest to Ours ($300$) for the other two seeds, as fixed before the runs; strength $30$ gives general $56.5$ and Hard $58.1$. Neither control reaches the keep loss on general R@1 or on any time-sensitive column. %

\subsection{SSv2-Label}
\label{sec:supp_keep_label}
SSv2-Label uses the $2{,}088$ SSv2-T videos with $1{,}989$ object-specific queries whose strings differ from the training templates, scored as SSv2-T (R@1, multiple positives). On Qwen, Plain training stays near the base model ($47.2$ with LoRA and $46.3$ with all weights, against $45.3$), although Plain training has a high SSv2-T ($63.2$ / $64.8$; with LoRA, the highest in \cref{tab:supp_abl_full}). Ours reaches $54.2$ and Ours (all weights) $59.1$; Ours (diff loss) gives $54.7$ with LoRA and $58.5$ with all weights, Ours w/o pair loss already $54.7$, and the diff loss at $w_p=5$ $53.0$. On InternVL the differences are small: $34.2$ for Plain (LoRA), $34.0$ for Ours, $35.2$ for Ours (diff loss) and $35.7$ for Ours w/o pair loss, against $23.8$ for the base model.  %

\section{Where the Gain Comes From}
\label{sec:supp_why}
This section adds evidence to \cref{sec:analysis} and to the paragraph ``Placement'' of \cref{sec:parts}: tests with changed frames, further reversal numbers, the placement numbers, the gain under cross-frame blocking, and the geometry of the video direction.

\subsection{Changed frames}
\label{sec:supp_why_frames}
We also change the frames of each test clip and evaluate again (three seeds). With the frames shuffled, both the base model and Ours drop, and Ours falls to the level of the base model (CiA-Charades / CiA-EPIC Chiral $52.0$ / $57.3$ for Ours against $52.5$ / $56.7$ for the base model), so the CiA gain of Ours over the base model vanishes (CiA-Charades $-0.5$, CiA-EPIC $+0.6$, against $+10.7$ and $+12.6$ on intact clips); this gain needs the order of the frames. Ours (diff loss) behaves the same ($52.1$ / $56.7$; gain $-0.5$ and $-0.1$, against $+8.2$ and $+10.1$ intact).
For Ours (diff loss) we also feed only the first or the second half of each clip, with each frame repeated twice. With the first half, the gain over the base model is $+5.4$ on CiA-Charades and $+6.8$ on CiA-EPIC; with the second half, it is $+1.2$ and $+4.8$. Both halves keep part of the gain on intact clips, and the first half keeps more. So the gains depend on frame order, not mainly on the end state of an action.

\subsection{Reversal: further numbers}
\label{sec:supp_why_reversal}
Three seeds each, original / reversed test clips: Ours w/o pair loss CiA-Charades $67.3$ / $42.7$, CiA-EPIC $69.4$ / $48.4$; the diff loss at $w_p=5$ $73.4$ / $39.5$ and $76.7$ / $45.8$; $\lambda\times5$ $71.2$ / $40.8$ and $73.4$ / $47.0$ (order shares $76$--$85\%$).
Qwen, base model $\to$ Ours: SSv2-T moves from $22.4$ to $51.9$ on the original clips and from $14.4$ to $26.8$ on the reversed clips (order share $29\%$).
InternVL, base model $\to$ Ours: CiA-Charades $54.8\to65.3$ (original) and $50.1\to43.2$ (reversed), CiA-EPIC $59.1\to68.8$ and $53.3\to47.6$ (order shares $83\%$ and $79\%$; SSv2-T $28\%$). With the diff loss: $84\%$, $78\%$, $27\%$.

\subsection{Placement}
\label{sec:supp_why_place}
This section gives the placement numbers behind the paragraph ``Placement'' of \cref{sec:parts}. With the pair-margin loss, the middle layers are above late-only LoRA on general R@1 ($+0.6$ $[0.2, 1.0]$), CiA-Charades ($+1.3$ $[0.8, 1.8]$) and CiA-EPIC ($+2.4$ $[1.1, 3.5]$), while RTime Hard differs little ($63.2$ vs.\ $62.1$). LoRA on all layers performs about the same as on the middle layers (RTime Hard $63.8$ vs.\ $63.2$). With the diff loss, other $12$-layer ranges that include the middle layers gain about as much as the middle layers (\cref{fig:placement}).

\subsection{Gain under cross-frame blocking}
\label{sec:supp_why_block}
\begin{figure}[h]
  \centering
  \includegraphics[width=.72\linewidth]{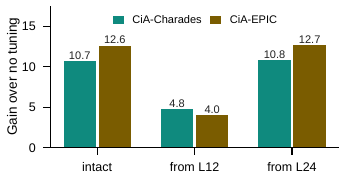}
  \caption{Gain of Ours over the base model in CiA Chiral mAP on the unseen benchmarks, with cross-frame attention blocked from L12 or L24 at test time in both the base model and Ours (three seeds).}
  \label{fig:supp_layers}
\end{figure}
We block cross-frame attention at test time in both the base model and Ours, and compare the gain of Ours over the base model (\cref{fig:supp_layers}). Blocking from L12 reduces the gain from $+10.7$ (CiA-Charades) and $+12.6$ (CiA-EPIC) to $+4.8$ and $+4.0$ (it keeps $44\%$ and $32\%$). Blocking from L24 leaves the gain at $+10.8$ and $+12.7$. So most of the gain of Ours (LoRA on L12--23) on these unseen benchmarks uses cross-frame composition in L12--L23. The adapters were trained on intact inputs, so part of this drop under blocking from L12 may come from the changed input. Late-only LoRA also gains less, on both Qwen and InternVL (\cref{fig:placement}). These results support the observation behind H3 (frames are combined in the middle layers); LoRA on the middle third of the layers gains about as much as LoRA on all layers (\cref{fig:placement}).  %

The same blocking test on Ours (all weights; three seeds): blocking from L12 keeps $86\%$ and $75\%$ of the gain on CiA-Charades and CiA-EPIC ($+9.1$ of $+10.6$ and $+9.4$ of $+12.5$; Ours (diff loss, all weights): $83\%$ and $76\%$), and blocking from L24 keeps all of the gain. So when all layers are trained, most of the gain no longer needs cross-frame attention after L12, and frames can be combined earlier. The middle-layer claim of the main text is about Ours with LoRA on L12--23. %
On InternVL (Ours, LoRA on L8--19, three seeds), blocking from L8 keeps $55\%$ and $50\%$ of the gain ($+5.8$ of $+10.6$ and $+4.9$ of $+9.7$), and blocking from L20 keeps all of the gain ($101\%$ and $98\%$), the same picture as on Qwen, although blocking from L8 removes a smaller share of the gain than blocking from L12 on Qwen.

\subsection{Agreement of differences}
\label{sec:supp_why_agree}
\begin{table}[t]
  \caption{Video direction and alignment on RTime test pairs (Qwen, mean over three seeds). $\lVert d_v\rVert$: length of the video direction (the language direction has length $0.67$ in the base model and $0.73$ for Ours); $\cos(d_v,d_t)$: alignment. Pair acc.: share of forward--reversed clip pairs that are told apart in both directions, \ie the forward query prefers the forward clip and the reversed query prefers the reversed clip. Pair margin: the pair-margin loss. RTime Hard follows the length of the video direction more than the alignment. Bold: highest in each column. Here and in \cref{sec:observations}, embeddings are L2-normalized, centered per modality over the forward and reversed items, and normalized again; without centering, the base model has the same alignment ($0.089$) and a video direction of length $0.27$ instead of $0.29$.}  %
  \label{tab:dir}
  \centering\footnotesize\setlength{\tabcolsep}{3pt}
  \begin{tabular}{@{}lcccc@{}}
    \toprule
    & $\lVert d_v\rVert$ & $\cos(d_v,d_t)$ & Pair acc. & Hard \\
    \midrule
    No tuning & 0.29 & 0.09 & 0.32 & 55.5 \\
    Ours w/o pair loss & 0.34 & 0.13 & 0.38 & 58.6 \\
    w/o pair loss, $\lambda{\times}5$ & 0.40 & 0.18 & 0.43 & 60.0 \\
    Ours (diff loss), $w_p{=}1$ & 0.37 & 0.18 & 0.44 & 60.7 \\
    Diff loss, $w_p{=}5$ & 0.44 & \textbf{0.27} & 0.51 & 62.1 \\
    \rowcolor{oursbg}\textbf{Ours} (pair margin), $w_p{=}3$ & 0.45 & 0.21 & 0.50 & 63.2 \\
    Pair margin, $w_p{=}5$ & \textbf{0.48} & 0.22 & \textbf{0.52} & \textbf{64.1} \\
    \bottomrule
  \end{tabular}
\end{table}
\begin{figure}[h]
  \centering
  \includegraphics[width=\linewidth]{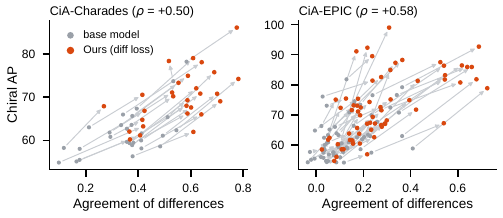}
  \caption{Pairs of opposite actions on the unseen CiA benchmarks, from the base model (gray) to Ours (diff loss) (orange; mean over three seeds). Agreement: cosine between the class-mean video difference and the text difference of the pair. Pairs whose differences come to agree more gain more Chiral AP ($\rho$: Spearman correlation between the change in agreement and the change in Chiral AP; $28$ pairs on CiA-Charades and $66$ on CiA-EPIC).}
  \label{fig:supp_agree}
\end{figure}
\paragraph{Length and alignment} The pair loss lengthens the video direction and turns it slightly toward the language direction, and RTime Hard follows this length more than the alignment (\cref{tab:dir}). The pair-margin loss, which aligns the two directions only implicitly, reaches a longer video direction and even a higher alignment ($0.21$ vs.\ $0.18$) than the diff loss at its default weight. At the same weight $w_p=5$, the diff loss aligns more ($0.27$ vs.\ $0.22$) but lengthens less and gains less on Hard ($62.1$ against $64.1$). The pair accuracy, the share of pairs told apart in both directions, rises from $0.32$ to $0.50$ with Ours. So the gain comes with widening each forward clip's gap to its reversed copy along its pair's language direction.

In \cref{tab:dir}, $\lambda\times5$ is the one exception to the link between length and RTime Hard: its video direction is long ($0.40$) but its RTime Hard is low ($60.0$). The diff loss optimizes the alignment directly, so for the diff loss a high alignment only shows that training reached its target on test data. The language direction of Ours (diff loss) keeps a cosine of $0.97$ to the language direction of the base model. On InternVL, Ours (diff loss) behaves as on Qwen (language norm $0.697\to0.728$, video norm $0.143\to0.252$, alignment $0.067\to0.160$; Ours w/o pair loss: $0.115$).
Agreement, in \cref{fig:supp_agree} (Ours (diff loss)), is a different quantity from the alignment: it compares class-mean differences of CiA pairs, not single RTime pairs. Within a benchmark, pairs whose class-mean video difference agrees more with the text difference have higher Chiral AP (Spearman about $0.5$--$0.7$ in every model; $0.47$ for the base model on CiA-Charades), and pairs whose agreement rises more gain more (about $0.5$). Both correlations are also positive for Ours w/o pair loss, $\lambda\times5$ and the pair-margin loss at $w_p=5$. Part of the link between agreement and Chiral AP is built into the scoring, since the dot product of the video difference with the text difference is the mean score gap. %

\subsection{Not a separate subspace}
\label{sec:supp_why_subspace}
We also ask whether training keeps general retrieval by moving the video direction out of the content subspace. In the top-$64$ principal components of the MSR-VTT test clip embeddings of the base model, the share of the video direction on RTime test pairs stays between $0.105$ and $0.141$ in every run we measure (Plain (LoRA) and variants of Ours (diff loss)), against $0.132$ in the base model. So general retrieval seems to be kept by holding the output embeddings near those of the base model (\cref{sec:supp_keep_neigh}), not by separating time from content. %

\section{Layer Analysis Details}
\label{sec:supp_layers}
These details complete the paragraph ``Frames are combined in the middle layers'' of \cref{sec:observations}.

\paragraph{Per-layer retrieval heads} In \cref{fig:layers_obs}a, b, on Qwen, the per-layer heads rise from about L17, and SSv2-T reaches $56.5$ at L28 ($44.6$ at the output) and RTime Hard $42.6$ at L29. On InternVL, which has fewer layers, SSv2-T peaks earlier, at $43.3$ at L15. The MSR-VTT head rises in the same layers, so the rise from about L17 is not specific to time.
Each head is a rank-$256$ residual map $x\mapsto\mathrm{norm}(x+xUV)$, $V=0$ at the start, from the last-token state of one layer of the frozen model into the final text embedding space of the frozen model, trained with InfoNCE. The RTime head is trained on RTime training clips (forward and reversed clips with their forward and reversed captions, the clip of the opposite direction and a frame-shuffled clip as extra negatives; $1{,}500$ steps; three seeds). The SSv2-T head is trained on SSv2 training clips, $50$ per template, against their label captions and the $174$ template names (three seeds). The MSR-VTT head is trained on the $9{,}000$ MSR-VTT training videos against one of their captions (one run). The number of steps of the SSv2-T and MSR-VTT heads is chosen on a $15\%$ hold-out of the training clips.

\paragraph{Linear probes} In \cref{fig:layers_obs}c, the appearance control is $79.3$ at L25 and $79.0$ at the output; opposite actions (CiA-SSv2) peak at L25 ($84.8$) and reach $77.6$ at the output. At the output, the retrieval heads give SSv2-T $44.6$ and RTime Hard $38.6$, and MSR-VTT $56.1$ (L35).
The probes are linear and trained per layer on centered, L2-normalized last-token states of the frozen model, with full-batch Adam ($300$ steps) and weight decay chosen from $\{10^{-5},10^{-4},10^{-3}\}$ on a $15\%$ hold-out. Opposite actions: a $32$-way softmax over the CiA-SSv2 classes, trained on $3{,}000$ CiA-SSv2 training clips and tested on the $1{,}430$ test clips; a clip counts as correct when its class scores above the opposite class (chance $50$). Appearance control: the same clips, $16$-way over the pairs with both directions merged (chance $6.25$). Forward versus reversed: a logistic probe on the difference between a forward RTime clip and its reversed copy, trained on $2{,}000$ RTime training clips (chance $50$). Every layer and condition uses the same training and test clips.

\paragraph{Blocking} On Qwen (\cref{sec:observations}), blocking from L12 lowers RTime Hard by $9.1$, CiA-SSv2 Chiral by $10.6$, CiA-SSv2 Static by $8.7$, and RTime Origin and MSR-VTT by $3.9$ each; blocking from L24 changes each score by at most $0.5$. Blocking from L12 also lowers the CiA-SSv2 probe at the output to $64.9$, while blocking from L24 leaves it at $77.1$.
On InternVL, blocking from L8 lowers RTime Hard by $2.9$, CiA-SSv2 Chiral by $3.5$ and SSv2-T by $7.5$, and also CiA-SSv2 Static by $3.7$, MSR-VTT by $3.1$ and RTime Origin by $2.4$; blocking from L20 changes each score by at most $1.1$ (RTime Hard $-0.8$).

\paragraph{Forward versus reversed} The forward-versus-reversed probe on RTime behaves differently from the CiA-SSv2 probe. This probe reaches $75.1$ already at the first layer and $82.2$ at L15, falls to $65.6$ at the output, and does not need cross-frame attention ($66.0$ at the output with blocking from L12). So the order of frames can fade before the output, as reported for question answering~\cite{shin2026fades}.  %
A linear probe on the text separates opposite actions already at the first layer ($99.8$), because the words of opposite actions differ.

\section{Full CiA Galleries}
\label{sec:supp_cia}
This section gives all three CiA galleries behind the Chiral columns of \cref{tab:main} and the wider-gallery results (Static and All) in \cref{sec:experiments}.
\begin{table*}[t]
  \caption{All CiA galleries on Qwen, mAP in percent (mean$\pm$std over three seeds). Chiral: the query class and its opposite. Static: the query class and one non-opposite class from every other pair. All: every class. Ours uses the pair-margin loss; the rows from Diff loss, $w_p{=}5$ down are variants of Ours (diff loss), except the two rows without the pair loss ($\lambda{\times}5$ and $\lambda{\times}10$). $\lambda{\times}k$: $k$ times the weight of the hard-negative margin loss. $^\ast$Unseen in training.} %
  \label{tab:cia_full}
  \centering\small\setlength{\tabcolsep}{3pt}
  \begin{tabular}{@{}lccc|ccc|ccc@{}}
    \toprule
 & \multicolumn{3}{c|}{CiA-Charades$^\ast$} & \multicolumn{3}{c|}{CiA-EPIC$^\ast$} & \multicolumn{3}{c}{CiA-SSv2} \\
 & Chiral & Static & All & Chiral & Static & All & Chiral & Static & All \\
    \midrule
    No tuning & 61.7 & 29.8 & 19.8 & 62.7 & 33.3 & 23.1 & 72.2 & 55.0 & 41.9 \\
    Plain (LoRA) & 71.4{\scriptsize$\pm$0.7} & 22.0{\scriptsize$\pm$0.6} & 17.5{\scriptsize$\pm$0.5} & 75.6{\scriptsize$\pm$1.1} & 25.4{\scriptsize$\pm$1.4} & 20.7{\scriptsize$\pm$0.9} & 87.7{\scriptsize$\pm$0.7} & 76.1{\scriptsize$\pm$2.1} & 66.2{\scriptsize$\pm$2.4} \\
    Ours w/o pair loss & 67.3{\scriptsize$\pm$0.1} & 27.6{\scriptsize$\pm$0.2} & 20.5{\scriptsize$\pm$0.1} & 69.4{\scriptsize$\pm$0.1} & 34.1{\scriptsize$\pm$0.8} & 26.1{\scriptsize$\pm$0.4} & 85.0{\scriptsize$\pm$0.5} & 75.7{\scriptsize$\pm$0.8} & 64.8{\scriptsize$\pm$0.8} \\
    \rowcolor{oursbg}\textbf{Ours} & 72.4{\scriptsize$\pm$0.2} & 24.7{\scriptsize$\pm$0.3} & 19.6{\scriptsize$\pm$0.2} & 75.3{\scriptsize$\pm$0.3} & 32.1{\scriptsize$\pm$1.2} & 26.2{\scriptsize$\pm$1.0} & 86.6{\scriptsize$\pm$0.8} & 72.8{\scriptsize$\pm$0.9} & 63.2{\scriptsize$\pm$0.6} \\
    Ours (diff loss) & 70.0{\scriptsize$\pm$0.1} & 26.1{\scriptsize$\pm$0.3} & 20.2{\scriptsize$\pm$0.2} & 72.8{\scriptsize$\pm$0.3} & 32.8{\scriptsize$\pm$1.0} & 26.3{\scriptsize$\pm$0.6} & 86.1{\scriptsize$\pm$0.2} & 74.3{\scriptsize$\pm$1.0} & 64.0{\scriptsize$\pm$0.8} \\
    \rowcolor{oursbg}\textbf{Ours} (all weights) & 72.3{\scriptsize$\pm$0.5} & 23.7{\scriptsize$\pm$0.5} & 18.8{\scriptsize$\pm$0.5} & 75.2{\scriptsize$\pm$1.1} & 26.7{\scriptsize$\pm$2.5} & 21.6{\scriptsize$\pm$2.1} & 88.9{\scriptsize$\pm$0.5} & 79.5{\scriptsize$\pm$0.9} & 69.4{\scriptsize$\pm$0.9} \\
    Ours (diff loss, all weights) & 71.2{\scriptsize$\pm$1.5} & 25.3{\scriptsize$\pm$0.3} & 19.8{\scriptsize$\pm$0.6} & 73.9{\scriptsize$\pm$0.9} & 28.6{\scriptsize$\pm$0.8} & 22.7{\scriptsize$\pm$0.7} & 88.4{\scriptsize$\pm$0.3} & 80.2{\scriptsize$\pm$0.6} & 69.1{\scriptsize$\pm$0.4} \\
    Plain (all weights) & 70.3{\scriptsize$\pm$1.2} & 18.7{\scriptsize$\pm$1.1} & 14.7{\scriptsize$\pm$1.0} & 72.6{\scriptsize$\pm$1.9} & 14.9{\scriptsize$\pm$0.2} & 11.7{\scriptsize$\pm$0.1} & 88.6{\scriptsize$\pm$0.4} & 78.3{\scriptsize$\pm$1.5} & 68.8{\scriptsize$\pm$2.0} \\
    Pair margin, $w_p{=}5$ & 73.3{\scriptsize$\pm$0.2} & 23.9{\scriptsize$\pm$0.3} & 19.1{\scriptsize$\pm$0.2} & 76.4{\scriptsize$\pm$0.5} & 31.5{\scriptsize$\pm$0.8} & 26.0{\scriptsize$\pm$0.7} & 86.8{\scriptsize$\pm$0.8} & 71.5{\scriptsize$\pm$0.6} & 62.1{\scriptsize$\pm$0.3} \\
    Pair margin + diff loss & 73.1{\scriptsize$\pm$0.3} & 23.9{\scriptsize$\pm$0.3} & 19.0{\scriptsize$\pm$0.2} & 76.4{\scriptsize$\pm$0.3} & 31.3{\scriptsize$\pm$1.1} & 26.1{\scriptsize$\pm$0.9} & 86.9{\scriptsize$\pm$0.9} & 72.1{\scriptsize$\pm$0.8} & 62.5{\scriptsize$\pm$0.5} \\
    Diff loss, $w_p{=}5$ & 73.4{\scriptsize$\pm$0.2} & 21.5{\scriptsize$\pm$0.2} & 17.1{\scriptsize$\pm$0.1} & 76.7{\scriptsize$\pm$0.4} & 27.4{\scriptsize$\pm$0.9} & 23.3{\scriptsize$\pm$0.7} & 87.3{\scriptsize$\pm$0.6} & 67.7{\scriptsize$\pm$0.7} & 58.4{\scriptsize$\pm$0.2} \\
    w/o pair loss, $\lambda{\times}5$ & 71.2{\scriptsize$\pm$0.3} & 25.3{\scriptsize$\pm$0.1} & 19.8{\scriptsize$\pm$0.0} & 73.4{\scriptsize$\pm$0.5} & 32.1{\scriptsize$\pm$0.7} & 25.6{\scriptsize$\pm$0.8} & 87.1{\scriptsize$\pm$0.6} & 73.7{\scriptsize$\pm$1.0} & 64.0{\scriptsize$\pm$1.0} \\
    w/o pair loss, $\lambda{\times}10$ & 72.9{\scriptsize$\pm$0.4} & 23.7{\scriptsize$\pm$0.2} & 18.9{\scriptsize$\pm$0.1} & 75.1{\scriptsize$\pm$0.7} & 30.7{\scriptsize$\pm$0.5} & 24.9{\scriptsize$\pm$0.6} & 87.6{\scriptsize$\pm$0.7} & 72.0{\scriptsize$\pm$1.3} & 62.5{\scriptsize$\pm$1.4} \\
    $\beta=3$ & 72.9{\scriptsize$\pm$0.2} & 23.3{\scriptsize$\pm$0.4} & 18.5{\scriptsize$\pm$0.3} & 76.8{\scriptsize$\pm$1.2} & 29.6{\scriptsize$\pm$1.0} & 24.6{\scriptsize$\pm$0.8} & 87.5{\scriptsize$\pm$0.4} & 73.7{\scriptsize$\pm$1.3} & 63.6{\scriptsize$\pm$1.4} \\
    Generated reversed queries & 71.5{\scriptsize$\pm$0.2} & 25.3{\scriptsize$\pm$0.2} & 19.8{\scriptsize$\pm$0.2} & 75.5{\scriptsize$\pm$0.3} & 32.8{\scriptsize$\pm$1.0} & 26.8{\scriptsize$\pm$0.6} & 87.7{\scriptsize$\pm$0.4} & 75.2{\scriptsize$\pm$1.0} & 65.4{\scriptsize$\pm$1.0} \\
    Diff loss, $w_p{=}5$, shuffled rev.\ query & 70.2{\scriptsize$\pm$0.3} & 24.9{\scriptsize$\pm$0.2} & 19.3{\scriptsize$\pm$0.1} & 72.3{\scriptsize$\pm$0.3} & 31.1{\scriptsize$\pm$1.1} & 25.3{\scriptsize$\pm$0.9} & 86.0{\scriptsize$\pm$0.6} & 72.7{\scriptsize$\pm$0.9} & 62.4{\scriptsize$\pm$0.9} \\
    Shuffled target & 67.4{\scriptsize$\pm$0.1} & 27.7{\scriptsize$\pm$0.2} & 20.5{\scriptsize$\pm$0.2} & 69.6{\scriptsize$\pm$0.2} & 34.0{\scriptsize$\pm$1.0} & 26.2{\scriptsize$\pm$0.6} & 84.9{\scriptsize$\pm$0.4} & 75.6{\scriptsize$\pm$0.7} & 64.8{\scriptsize$\pm$0.7} \\
    \bottomrule
  \end{tabular}
\end{table*}
\begin{table*}[t]
  \caption{All CiA galleries on InternVL, mAP in percent (mean$\pm$std over three seeds). Galleries as in \cref{tab:cia_full}. Ours uses the pair-margin loss with $w_p{=}5$. $^\ast$Unseen in training.}
  \label{tab:cia_full_ivl}
  \centering\small\setlength{\tabcolsep}{3pt}
  \begin{tabular}{@{}lccc|ccc|ccc@{}}
    \toprule
 & \multicolumn{3}{c|}{CiA-Charades$^\ast$} & \multicolumn{3}{c|}{CiA-EPIC$^\ast$} & \multicolumn{3}{c}{CiA-SSv2} \\
 & Chiral & Static & All & Chiral & Static & All & Chiral & Static & All \\
    \midrule
    No tuning & 54.8 & 23.0 & 12.7 & 59.1 & 19.1 & 11.9 & 61.1 & 40.4 & 26.3 \\
    Plain (LoRA) & 62.5{\scriptsize$\pm$0.7} & 28.2{\scriptsize$\pm$0.5} & 19.0{\scriptsize$\pm$0.2} & 65.9{\scriptsize$\pm$1.5} & 19.2{\scriptsize$\pm$0.8} & 14.0{\scriptsize$\pm$0.4} & 77.9{\scriptsize$\pm$0.7} & 67.1{\scriptsize$\pm$1.0} & 52.7{\scriptsize$\pm$0.4} \\
    Ours w/o pair loss & 59.6{\scriptsize$\pm$0.3} & 27.3{\scriptsize$\pm$0.1} & 17.2{\scriptsize$\pm$0.0} & 64.5{\scriptsize$\pm$0.8} & 21.3{\scriptsize$\pm$0.2} & 14.8{\scriptsize$\pm$0.2} & 74.0{\scriptsize$\pm$0.5} & 64.1{\scriptsize$\pm$0.4} & 49.6{\scriptsize$\pm$0.1} \\
    \rowcolor{oursbg}\textbf{Ours} & 65.3{\scriptsize$\pm$0.1} & 25.6{\scriptsize$\pm$0.3} & 17.8{\scriptsize$\pm$0.2} & 68.8{\scriptsize$\pm$0.4} & 18.6{\scriptsize$\pm$0.3} & 13.5{\scriptsize$\pm$0.2} & 78.2{\scriptsize$\pm$0.4} & 62.4{\scriptsize$\pm$1.5} & 50.1{\scriptsize$\pm$1.0} \\
    Ours (diff loss) & 61.7{\scriptsize$\pm$0.1} & 26.7{\scriptsize$\pm$0.2} & 17.4{\scriptsize$\pm$0.1} & 66.4{\scriptsize$\pm$0.9} & 20.2{\scriptsize$\pm$0.5} & 14.6{\scriptsize$\pm$0.3} & 76.4{\scriptsize$\pm$0.7} & 63.8{\scriptsize$\pm$0.6} & 50.8{\scriptsize$\pm$0.3} \\
    Diff loss, $w_p{=}5$ & 65.1{\scriptsize$\pm$0.4} & 23.8{\scriptsize$\pm$0.2} & 16.4{\scriptsize$\pm$0.2} & 69.6{\scriptsize$\pm$0.9} & 17.2{\scriptsize$\pm$0.1} & 13.0{\scriptsize$\pm$0.2} & 78.3{\scriptsize$\pm$0.1} & 60.7{\scriptsize$\pm$0.9} & 49.2{\scriptsize$\pm$0.5} \\
    w/o pair loss, $\lambda{\times}5$ & 62.2{\scriptsize$\pm$0.5} & 26.3{\scriptsize$\pm$0.2} & 17.3{\scriptsize$\pm$0.0} & 67.0{\scriptsize$\pm$1.0} & 20.0{\scriptsize$\pm$0.3} & 14.3{\scriptsize$\pm$0.1} & 77.6{\scriptsize$\pm$0.9} & 63.6{\scriptsize$\pm$0.8} & 50.9{\scriptsize$\pm$0.3} \\
    \bottomrule
  \end{tabular}
\end{table*}

\Cref{tab:cia_full} gives all three CiA galleries on Qwen and \cref{tab:cia_full_ivl} on InternVL. The main text reports Chiral (the query class against its opposite); Static compares the query class with one non-opposite class from every other pair, and All with every class (\cref{sec:observations}).

\paragraph{Static and All} Against the base model, Ours on Qwen changes Static by $-5.1$ and $-1.1$ and All by $-0.2$ and $+3.1$ (CiA-Charades and CiA-EPIC). Ours (diff loss) loses less on Static than Ours ($-3.7$ and $-0.4$ on CiA-Charades and CiA-EPIC). In the ablations of Ours (diff loss), the Static drop comes mainly from the output heads (\cref{sec:supp_abl_parts}). On InternVL, Ours raises CiA-Charades Static and All over the base model ($+2.6$ and $+5.1$) and keeps CiA-EPIC Static ($-0.5$) (\cref{tab:supp_ci}).  %

\paragraph{Object split} For Ours, Static falls mainly on classes with the same verb and another object (CiA-Charades $-10.4$ $[-12.9, -8.1]$, CiA-EPIC $-7.4$ $[-9.5, -5.2]$ against the base model; Ours (diff loss): $-8.5$ and $-5.8$).

\section{Unseen Event Order: Vinoground}
\label{sec:supp_vino}
This section gives the full Vinoground results behind the sentence on unseen event order in \cref{sec:experiments}.
Vinoground~\cite{zhang2024vinoground} has $500$ pairs of short clips and captions. The two captions of a pair use the same words with the events in the other order, and the two clips show the events in the two caption orders. We score Vinoground zero-shot with the retrieval embedding of each model (\cref{sec:experiments}). The text score counts a pair as correct when both clips prefer their own caption, the video score when both captions prefer their own clip, and the group score when both the text and the video conditions hold (chance $25$, $25$ and $16.7$).
\Cref{tab:supp_vino} gives the results. Ours is above the base model on all three scores (text $+4.9$ $[2.1, 7.9]$, video $+2.7$ $[0.2, 5.3]$, group $+2.4$ $[0.5, 4.3]$), and above Ours w/o pair loss (text $+3.9$ $[1.1, 6.7]$, video $+1.9$ $[0.2, 3.7]$, group $+1.7$ $[0.4, 3.1]$), so the gain of Ours over the base model comes mostly from the pair loss. Plain (LoRA) reaches about the same scores as Ours (group $6.5$) but loses $3.7$ general R@1 (\cref{tab:main}). The largest gain of Ours over the base model is on the object category. All models stay below chance, as embedding models do in the original study~\cite{zhang2024vinoground}. Inverting the choices would not raise these scores, since the text score counts a pair only if each clip prefers its own caption, and the video score only if each caption prefers its own clip.  %
\begin{table}[h]
  \caption{Vinoground (zero-shot, Qwen; mean$\pm$std over three seeds). Text, video, group: scores over all $500$ pairs. Right: group score by category (action $257$, object $160$, viewpoint (Viewp.) $83$ pairs). No model is trained on Vinoground.}
  \label{tab:supp_vino}
  \centering\footnotesize\setlength{\tabcolsep}{2pt}
  \begin{tabular}{@{}lccc|ccc@{}}
    \toprule
    & Text & Video & Group & Action & Object & Viewp. \\
    \midrule
    No tuning & 18.8 & 8.2 & 3.8 & 2.7 & 6.3 & 2.4 \\
    Plain (LoRA) & 22.8{\scriptsize$\pm$0.2} & 11.7{\scriptsize$\pm$0.8} & 6.5{\scriptsize$\pm$0.6} & 4.3 & 12.3 & 2.4 \\
    Ours w/o pair loss & 19.8{\scriptsize$\pm$1.7} & 9.0{\scriptsize$\pm$0.2} & 4.5{\scriptsize$\pm$0.5} & 2.1 & 9.6 & 2.0 \\
    Ours (diff loss) & 22.1{\scriptsize$\pm$0.4} & 9.7{\scriptsize$\pm$0.2} & 5.3{\scriptsize$\pm$0.3} & 2.5 & 11.0 & 2.8 \\
    w/o pair loss, $\lambda{\times}5$ & 22.9{\scriptsize$\pm$1.3} & 9.5{\scriptsize$\pm$0.6} & 5.3{\scriptsize$\pm$1.2} & 2.3 & 11.5 & 2.4 \\
    \rowcolor{oursbg}\textbf{Ours} & 23.7{\scriptsize$\pm$1.0} & 10.9{\scriptsize$\pm$0.3} & 6.2{\scriptsize$\pm$0.2} & 2.9 & 12.5 & 4.4 \\
    \bottomrule
  \end{tabular}
\end{table}

\section{Comparison Details}
\label{sec:supp_cmp}
This section gives details for the comparison in \cref{sec:experiments}: ensembles, the protocol for released models, and WiSE-FT.

\subsection{Ensembles}
\label{sec:supp_cmp_ens}
TARA also reports an ensemble of its released model with Qwen3-VL-Embedding, our base model: TARA concatenates the embeddings of the two models, which ranks like adding the two similarities with weight $0.5$; we keep this weight fixed. TARA $\oplus$ Qwen combines the released Tarsier2 + TARA model with the base model. TARA $\oplus$ Qwen is close to Ours, a single model, on RTime Hard ($63.5$ vs.\ $63.2$) and higher on CiA-EPIC ($76.2$ vs.\ $75.3$), at about the same general R@1. TARA $\oplus$ Qwen needs two embedding models and a vector twice as long for each clip, and it is lower on CiA-Charades ($67.7$ vs.\ $72.4$).
Ours is also a better ensemble partner for TARA than the base model. TARA $\oplus$ Ours keeps the general R@1 of TARA $\oplus$ Qwen ($58.0$ vs.\ $58.2$) and is above TARA $\oplus$ Qwen on every time-sensitive column: RTime Hard $66.2$, Binary $75.2$, CiA-Charades $72.6$, CiA-EPIC $80.5$ and SSv2-T $54.6$ (three seeds). TARA $\oplus$ Ours (diff loss) is also above TARA $\oplus$ Qwen on every time-sensitive column at the same general R@1 ($58.2$; RTime Hard $64.7$, Binary $73.7$, CiA-Charades $71.2$, CiA-EPIC $79.5$, SSv2-T $56.3$), and below TARA $\oplus$ Ours on every time-sensitive column except SSv2-T.

\subsection{Released models and the TARA rerun}
\label{sec:supp_cmp_tara}
We evaluate each released model with its own model code, prompts and pooling, on our $16$ frames, test subsets, query lists and metrics, so scores can differ from the original papers. Each embedding is L2-normalized.
\emph{Tarsier2 + TARA and Tarsier2.} The released TARA weights on Tarsier2-7B and zero-shot Tarsier2-7B use the TARA model class and prompts (``Summary above video in one word:'' and ``Summary above sentence in one word:''), with the last-layer state of the last prompt token; our $16$ frames are passed through the TARA video path. Zero-shot Tarsier2-7B has general R@1 $12.3$, RTime Hard $19.5$, Binary $53.7$, CiA-Charades $54.4$, CiA-EPIC $63.4$ and SSv2-T $22.4$, so TARA training raises its general R@1 from $12.3$ to $48.4$. In the zero-shot Tarsier2-7B run, $4$ CiA-Charades and $5$ CiA-EPIC clips could not be decoded (about $0.1\%$ and $0.2\%$ of each gallery); they stay in the gallery with a zero embedding.
\emph{CaRe-7B.} We use the encoder of the CaReBench code: the same one-word prompts, each frame duplicated so that one Qwen2-VL temporal patch covers one input frame, and the last-token state. The CaRe paper uses $32$ frames.
\emph{GVE-7B.} The video side follows the GVE model card (last-token state; the GVE paper uses $8$ frames). The card gives no query format, so we put the retrieval instruction of the GVE paper (Table 5 there) in the system turn and end each query with the end-of-text token; without this token the query embeddings collapse (MSR-VTT R@1 $5.6$).  %
\emph{LamRA-Ret.} We follow the official LamRA zero-shot code: ``Summarize above video in one word:'' for clips, and ``Find an everyday video match with caption.'' followed by the same summary prompt for queries (the official query prompt exists only for MSR-VTT and MSVD; we use it everywhere), with the state before the embedding token that LamRA adds. The official LamRA reader samples $2$ frames per second; we feed our $16$ frames.
Against the published numbers, CaRe-7B is within $2.3$ R@1 on MSR-VTT and DiDeMo. GVE-7B is $1.9$ lower on MSR-VTT and about $9$ lower on DiDeMo ($34.0$ vs.\ $43.3$); whole videos and $8$ frames do not close the DiDeMo gap.

The TARA code for Qwen3-VL-Embedding is not released, so we rerun the recipe from the released trainer (\cref{sec:supp_setup_tara}).
TARA reports MSR-VTT for Qwen3-VL-Embedding-8B as $79.8$ (its Table 11, ``Std''); this is R@5 on the NegBench version of MSR-VTT ($1{,}000$ queries)~\cite{alhamoud2025negbench}, not R@1. Under our protocol, the same base model has R@5 $79.3$ on MSR-VTT 1k-A (R@1 $56.1$, which we report throughout), and our rerun of the TARA LoRA setting has R@5 $77.3$ (mean of three seeds) against $78.8$ in its paper, so the two protocols agree. Our rerun of the full recipe lowers R@5 more ($64.5$ against $70.2$), in the same direction; the data and prompt TARA used on Qwen are not released. On the same model, TARA reports CiA-SSv2 Chiral $72.0$ before fine-tuning~\cite{bagad2026tara}, close to the $72.2$ of the base model in our protocol (\cref{tab:main}). %

\subsection{WiSE-FT}
\label{sec:supp_cmp_wise}
WiSE-FT~\cite{wortsman2022wise} interpolates the weights $\theta$ after Plain training with the base weights $\theta_0$: $\theta_0+\alpha(\theta-\theta_0)$. For LoRA we scale the LoRA update and the head update by $\alpha$, which is exact; for all weights we interpolate every trained weight. \Cref{tab:supp_wise} gives three seeds per point. WiSE-FT comparisons at equal or matched general R@1 interpolate the scores linearly between the two evaluated $\alpha$ that bracket it; the interpolation weight is set from the three-seed means and kept fixed in the bootstrap. With LoRA, the WiSE-FT curve of Plain training passes through Ours (diff loss) and stays below Ours. With all weights, the WiSE-FT curve of Plain training stays below Ours (all weights): at $\alpha=0.5$, its best point on RTime Hard, general R@1 is $0.8$ $[0.3, 1.5]$ lower, and so are Binary ($2.2$ $[0.5, 4.0]$) and CiA-Charades ($1.8$ $[0.8, 2.9]$); at equal general R@1 Ours (all weights) is higher on Binary ($+3.0$ $[1.4, 4.6]$), CiA-Charades ($+2.9$ $[1.9, 3.9]$) and SSv2-T ($+7.8$ $[1.4, 14.1]$), and not significantly different on Hard and CiA-EPIC. The WiSE-FT curve of Plain training also stays, at matched general R@1, below Ours (diff loss, all weights) on all columns but CiA-EPIC. %

We also train the pair-margin loss without the keep loss ($\beta=0$; general $54.7$) and apply WiSE-FT to this model ($\alpha=0.25$, $0.5$, $0.75$; three seeds). At the general R@1 of Ours, this WiSE-FT curve reaches about the same point as Ours: Ours differs by $+0.4$ $[-1.0, 1.7]$ on RTime Hard, $+0.7$ $[-0.5, 1.9]$ on Binary, $+0.3$ $[-0.3, 0.8]$ and $+0.9$ $[0.0, 1.8]$ on CiA-Charades and CiA-EPIC, and $-3.3$ $[-9.3, 2.8]$ on SSv2-T. Both hold the output embeddings near those of the base model (closeness $0.984$ for Ours, $0.94$ and $0.88$ at $\alpha=0.5$ and $0.75$), and general R@1 follows closeness in both (\cref{sec:supp_keep_neigh}). So the keep loss reaches, in one training run and without choosing $\alpha$, the point that post-hoc interpolation of a no-keep model reaches, and the gain of Ours over WiSE-FT of Plain training comes from the pair loss.
\begin{table}[h]
  \caption{WiSE-FT on Qwen (mean over three seeds): $\theta_0+\alpha(\theta-\theta_0)$ between the base model and Plain. General: mean R@1 over the six general benchmarks. Hard and Binary: RTime. Cha., EPIC: CiA-Charades and CiA-EPIC Chiral. All w.: all weights. $^\ast$Unseen in training.}
  \label{tab:supp_wise}
  \centering\footnotesize\setlength{\tabcolsep}{3pt}
  \begin{tabular}{@{}lcccccc@{}}
    \toprule
    & General & Hard & Binary & Cha.$^\ast$ & EPIC$^\ast$ & SSv2-T \\
    \midrule
    No tuning & 58.4 & 55.5 & 62.8 & 61.7 & 62.7 & 22.4 \\
    \multicolumn{7}{@{}l}{\emph{WiSE-FT of Plain (LoRA)}} \\
    $\alpha=0.25$ & 59.1 & 59.2 & 66.2 & 65.9 & 66.4 & 38.5 \\
    $\alpha=0.5$ & 58.9 & 61.0 & 68.5 & 68.6 & 70.2 & 53.4 \\
    $\alpha=0.75$ & 57.5 & 60.5 & 69.2 & 70.4 & 73.4 & 60.2 \\
    Plain, $\alpha=1$ & 54.7 & 57.8 & 68.7 & 71.4 & 75.6 & 63.2 \\
    \textbf{Ours} & 58.4 & 63.2 & 72.0 & 72.4 & 75.3 & 51.9 \\
    \textbf{Ours} (diff loss) & 58.3 & 60.7 & 69.3 & 70.0 & 72.8 & 54.0 \\
    \multicolumn{7}{@{}l}{\emph{WiSE-FT of Plain (all weights)}} \\
    $\alpha=0.25$ & 58.7 & 62.1 & 69.5 & 67.9 & 73.0 & 46.7 \\
    $\alpha=0.5$ & 56.7 & 63.4 & 71.3 & 70.5 & 76.1 & 63.4 \\
    $\alpha=0.75$ & 53.1 & 61.7 & 71.2 & 70.8 & 74.4 & 65.7 \\
    Plain, $\alpha=1$ & 50.1 & 59.7 & 70.7 & 70.3 & 72.6 & 64.8 \\
    \textbf{Ours} (all w.) & 57.5 & 64.2 & 73.5 & 72.3 & 75.2 & 64.4 \\
    \textbf{Ours} (diff loss, all w.) & 57.9 & 63.3 & 72.1 & 71.2 & 73.9 & 65.7 \\
    \bottomrule
  \end{tabular}
\end{table}

\end{document}